\documentclass[twoside]{article}

\usepackage[accepted]{aistats2024}
\usepackage[utf8]{inputenc} 
\usepackage[T1]{fontenc}    
\usepackage{xcolor}         
\definecolor{mydarkblue}{rgb}{0,0.08,0.45}
\usepackage[colorlinks, anchorcolor=white, linkcolor=blue, urlcolor=magenta, citecolor=mydarkblue]{hyperref}
\usepackage{url}            
\usepackage{wrapfig}
\usepackage{booktabs}       
\usepackage{amsfonts}       
\usepackage{nicefrac}       
\usepackage{microtype}      
\usepackage{xcolor}         

\usepackage{graphicx}
\usepackage{subfigure}

\usepackage{hyperref}
\usepackage{multirow}
\usepackage{algorithm}
\usepackage{algorithmic}

\usepackage{amsmath, amsfonts}
\usepackage{amssymb}
\usepackage{mathtools}
\usepackage{amsthm}
\usepackage{bbm}

\usepackage[capitalize,noabbrev]{cleveref}
\usepackage{enumitem}

\usepackage{tikz}
\usepackage{tcolorbox}
\newtcbox{\alertinline}[1][black]
  {on line, arc = 0pt, outer arc = 0pt,
    colback = #1!5!white, colframe = #1!50!black,
    boxsep = 0pt, left = 1pt, right = 1pt, top = 2pt, bottom = 2pt,
    boxrule = 0pt, bottomrule = 0.7pt, toprule = 0.7pt}

\theoremstyle{plain}
\newtheorem{theorem}{Theorem}[section]

\newtheorem{lemma}[theorem]{Lemma}

\theoremstyle{definition}
\newtheorem{definition}[theorem]{Definition}

\theoremstyle{remark}

\usepackage{makecell}
\usepackage{caption}

\newcommand{\FLT}{$\mathrm{FLT}$}
\newcommand{\FLTs}{$\mathrm{FLTs}$}

\begin{document}

%
\runningtitle{Learning a Fourier Transform for Linear Relative Positional Encodings in Transformers}

%
\runningauthor{Choromanski, Li, Likhosherstov, Dubey, Luo, He, Yang, Sarlos, Weingarten, Weller}

\twocolumn[

\aistatstitle{Learning a Fourier Transform for\\Linear Relative Positional Encodings in Transformers}

\aistatsauthor{Krzysztof Marcin Choromanski$^{*1,2}$ $\qquad$
Shanda Li$^{*3}$  $\qquad$
Valerii Likhosherstov$^{4}$ 
}
\aistatsauthor{Avinava Dubey$^{1}$  $\qquad$ Shengjie Luo$^{5}$  $\qquad$ Di He$^{5}$   $\qquad$  Yiming Yang$^{3}$ 
}
\aistatsauthor{Tamas Sarlos$^{1}$  $\qquad$
Thomas Weingarten$^{1}$  $\qquad$
Adrian Weller$^{6,7}$  }
\vspace{0.5mm}
\aistatsaddress{$^{1}$Google Research $\qquad$ $^{2}$Columbia University $\qquad$
$^{3}$Carnegie Mellon University}
\vspace{-7.5mm}
\aistatsaddress{$^{4}$Waymo  $\qquad$
$^{5}$Peking University  $\qquad$
$^{6}$University of Cambridge $\qquad$
$^{7}$Alan Turing Institute }
\vspace{-7.5mm}
\aistatsaddress{\texttt{\{kchoro,avinavadubey\}@google.com}  $\qquad$
\texttt{shandal@cs.cmu.edu}}
\vspace{-6mm}
\aistatsaddress{$^*$\footnotesize\sffamily These two authors contributed equally. The authorship is in alphabetical order.}
]

\begin{abstract}
We propose a new class of linear Transformers called $\mathrm{FourierLearner}$-$\mathrm{Transformers}$ (\FLTs), which incorporate a wide range of relative positional encoding mechanisms (RPEs). These include regular RPE techniques applied for sequential data, as well as novel RPEs operating on geometric data embedded in higher-dimensional Euclidean spaces. \FLTs\ construct the optimal RPE mechanism implicitly by learning its spectral representation. As opposed to other architectures combining efficient low-rank linear attention with RPEs, \FLTs\ remain practical in terms of their memory usage and do not require additional assumptions about the structure of the RPE mask. Besides, \FLTs\ allow for applying certain structural inductive bias techniques to specify masking strategies, e.g. they provide a way to learn the so-called \textit{local RPEs} introduced in this paper and give accuracy gains as compared with several other linear Transformers for language modeling. We also thoroughly test \FLTs\ on other data modalities and tasks, such as image classification, 3D molecular modeling, and learnable optimizers. To the best of our knowledge, for 3D molecular data, \FLTs\ are the first Transformer architectures providing linear attention and incorporating RPE masking.
\end{abstract}

\vspace{-2mm}
\section{INTRODUCTION}
\label{sec:intro}


Transformers have revolutionized the landscape of machine learning, introducing a paradigm shift in the way that people approach complex tasks in natural language processing (NLP) \cite{devlin}, computer vision (CV) \cite{vits}, molecular modeling \cite{jumper2021highly}, and beyond. 

The largest computational bottleneck in Transformers is also the source of their success, the attention module. The attention module propagates signals between different tokens in the input sequence and has quadratic time and space complexity with respect to the input length $L$, which limits its scalability to long sequences. Thus, designing efficient attention modules has been an active area of research. Recently, the research on ``efficient'' Transformers has taken on new importance as the size of Transformer models grew from the GPT-1 architecture of ``only'' \textbf{117M} parameters to GPT-3 with \textbf{175B} parameters, a \textbf{1000$\times$} increase within just two years \cite{gpt3}. 

One class of efficient Transformers is based on \textit{sparse attention} \cite{li2019enhancing, local-attention,zaheer2020big,routing, apoorv, reformer, sparse-hash}. 
These methods do not aim at approximating the regular attention, but rather propose simpler and more tractable attention mechanisms, sometimes with additional constraints (e.g. identical queries and keys \cite{reformer}). Another popular class of efficient Transformers is based on the \textit{kernelized attention} \cite{choromanski,tsai2019transformer,katharopoulos2020transformers}. The key idea is to find an approximate low-rank decomposition of the attention matrix and leverage it to improve space and time complexity of the attention mechanism via the associativity property of matrix multiplications. Performer \cite{choromanski} is a successful example of this model class. In contrast to previously discussed methods, Performer's approximate attention matrix (which is never explicitly constructed but rather implicitly used) is an unbiased estimate of the original attention matrix encoding similarities between tokens via the softmax kernel. Performers have been adopted into many Transformer stacks to provide linear space and time complexity \cite{perf-vit-1, perf-vit-2, tay2021omninet,performer-mpc}.

Unfortunately, the simplicity of Performers comes at a price. It is well known that incorporating structural inductive priors -- which is usually implemented via various additive relative masking mechanisms in regular attention architectures -- is difficult for Performers. We refer to these methods as \textit{Relative Positional Encodings} (RPEs) \cite{shaw,raffel,li2021can,luo2022your}.
RPEs play a critical role in improving the performance of Transformers in long-range modeling for language \cite{dai-etal-2019-transformer}, speech \cite{liutkus}, vision \cite{wupeng}, and genomic data \cite{Avsec2021EffectiveGE}. However, at first glance, Performers are not compatible with general RPE techniques, since they seem to require explicit materialization of the attention matrix to apply the RPE mask, which is exactly what Performers avoid in order to achieve computational improvements.
Substantial efforts are made to reconcile Performers with RPEs (more details in Sec. \ref{sec:related}), but so far all these attempts fall short of providing two properties at the same time: (a) practical computational gains, and (b) inclusion of general RPE methods, for inputs with nontrivial topological structures.

In this paper, we propose a new class of linear Transformers called $\mathrm{FourierLearner}$-$\mathrm{Transformers}$ (\FLTs), which incorporate a wide range of relative positional encoding mechanisms (RPEs). These include regular RPE techniques applied for sequential data, and novel RPEs operating on geometric data embedded in higher-dimensional Euclidean spaces (e.g. molecular structures). \FLTs\ construct the optimal RPE mechanism implicitly by learning its spectral representation, and enjoy provable uniform convergence guarantees. As opposed to other architectures combining efficient low-rank linear attention with RPEs, \FLTs\ remain practical in terms of their memory usage and do not require additional assumptions about the structure of the RPE mask. Besides, \FLTs\ allow the application of certain structural inductive bias techniques to specify masking strategies, e.g. they provide a way to learn what we call \textit{local RPEs}, introduced in this paper and providing accuracy gains compared with several other linear Transformers for language modeling. We also thoroughly test \FLTs\ on other data modalities and tasks, such as image classification and molecular modeling. To the best of our knowledge, for 3D molecular data, \FLTs\ are the first Transformer architectures providing linear attention and incorporating RPE masks, which broadens the scope of RPE-enhanced linear attention.

To summarize, our main contributions are as follows:
\vspace{-2mm}
\begin{itemize}
    \item We introduce the proposed RPE-enhanced linear attention,    $\mathrm{FourierLearner}$-$\mathrm{Transformers}$ (\FLTs). \FLTs~are applicable to not only sequential data (e.g., texts) but also geometric data embedded in higher-dimensional Euclidean spaces (e.g., 3D molecular data), significantly broadening the scope of RPE-enhanced linear attention.
    
    \item We provided detailed theoretical analysis on \FLTs, including the uniform convergence and sample complexity bound on its approximation (Sec. \ref{subsec:flt}). We discuss several instantiations, in particular \FLTs\ with so-called Gaussian mixture RPEs, shift-invariant kernel RPEs and local RPEs (Sec. \ref{sec:topology}).

    \item We extensively evaluate \FLTs~on language modeling (Sec. \ref{sec:lm}), image classification (Sec. \ref{sec:images}), and molecular property predictions (Sec. \ref{sec:molecular_dynamics}). Our experiments show that \FLTs~can be easily applied to a wide range of data modalities and demonstrate strong performance and efficiency. 
\end{itemize}

\section{RELATED WORKS}
\label{sec:related}
\paragraph{Kernelized attention with RPE.}
One of the first attempts to address the problem of combining kernelized attention Transformers with RPEs is \cite{liutkus}, where two variants, namely $\mathrm{sineSPE}$ and $\mathrm{convSPE}$, are proposed. Both variants model the RPE mask as a stationary position kernel with a Toeplitz mask structure. While their complexity is linear in the sequence length $L$, extra dependency on the number of sinusoidal components $T$ (for $\mathrm{sineSPE}$) / the convolution filter lengths $P$ (for $\mathrm{convSPE}$) is introduced. In practice, $T$ or $P$ has to be sufficiently small due to computational budgets. Besides, they constrain the RPE mask to be a valid kernel matrix, while our \FLTs~do not require such assumptions. Both $\mathrm{sineSPE}$ and $\mathrm{convSPE}$ significantly underperform \FLTs~on language modeling (Sec. \ref{sec:lm}). And they cannot be applied for more general RPEs with tokens embedded in the higher-dimensional Euclidean spaces, e.g., RPEs for 3D molecular data.

Recently, \cite{rpe-performers, topmasking} show that the RPE mechanism can be combined with Performers in $O(L\log(L))$ time complexity. The method relies on the elegant observation that log-linear time complexity can be achieved as long as the exponentiated RPE mask supports fast matrix-vector multiplication. RPEs for sequential data satisfy this condition since the corresponding masks have a Toeplitz structure. However, this method has large space complexity and high memory consumption in practice (Sec. \ref{sec:lm}). 
Moreover, it heavily relies on the structure of sequential data and does not apply to 3D molecular data where the RPE masks do not have a Toeplitz structure.

\paragraph{Random Fourier features (RFFs).} There has been voluminous literature on the field of RFFs \cite{rahimi, kapralov-rfs, szabo, li-rfs, hybrid_rfs, crts, tr-kernel}. However, the research on learnable RFF variants \cite{learning-rfs-1} is relatively new. Furthermore, prior works are mostly narrowed to applying RFFs in the context of positive definite (PD) kernels, while our work breaks this limitation since RPEs do not need to be defined by PD kernels. Several papers also explore the development of Transformer-based models whose attention mechanism operates in the spectral domain \cite{tamkin2020language, moreno2023deep}, but they do not study efficient RPE modeling.


\paragraph{Long sequence modeling.} Applying deep learning models to long sequences is an active research direction. 
We study efficient Transformer models for long sequence modeling. 
While our focus lies within the Transformer realm, it's worth noting the existence of alternative, non-Transformer architectures \cite{ham,bello2021lambdanetworks, s4}. 
Beyond efficiency, \cite{o2021context, liu2024lost} probe context usage of long sequence language models; \cite{press2022train, ruoss2023randomized, li2024functional} design sequence models with length generalization ability (i.e., training on short sequences and generalize to long sequences); \cite{yun2020n, yang2024efficient} study the theoretical capability of those models. 
Note that existing works most focus on one data modality, while \FLT~is evaluated across a wide range modalities.
\section{PRELIMINARIES}
\label{sec:preliminaries}
\paragraph{General RPE mechanism in Transformers.}
Consider an input sequence $\mathbf{X}\in\mathbb{R}^{L\times d_{\text{in}}}$ where $L$ and $d_{\text{in}}$ denote the number and embedding size of tokens. The self-attention module in Transformers linearly projects the input into three matrices $\mathbf{Q}, \mathbf{K}, \mathbf{V} \in \mathbb{R}^{L \times d}$ called \textit{queries}, \textit{keys} and \textit{values} respectively. We also associate all the tokens with positional features $\mathbf{r}_{1},...,\mathbf{r}_{L} \in \mathbb{R}^{\ell}$ that are used to define the relative positional encoding (RPE) mechanism below:

\begin{definition}[General RPE for attention]
\label{gen_graph_attention}
\textit{General Relative Positional Encoding} enhanced attention is of the following form, where $\mathbf{N} = [f(\mathbf{r}_{i}-\mathbf{r}_{j})]_{i,j\in [L]} \in \mathbb{R}^{L \times L}$ is the so-called \textit{RPE mask}\footnote{We use $[L]$ to denote $\{1, \cdots, L\}$ in this paper.} and $f:\mathbb{R}^{\ell} \rightarrow \mathbb{R}$ is a (potentially learnable) functon:
\vspace{-1.5mm}
\begin{align}
    &\mathrm{Att}(\mathbf{Q}, \mathbf{K}, \mathbf{V},\mathbf{N}) = \mathbf{D}^{-1} \mathbf{A} \mathbf{V}, \nonumber \\
    \text{where }& \mathbf{A} = \exp \left(\mathbf{N}  +\frac{\mathbf{Q} \mathbf{K}^\top} {\sqrt{d}}\right), \mathbf{D} = \mathrm{diag} ( \mathbf{A} \mathbf{1}_L ).  \label{eq:attnorm1}
\end{align}
\end{definition}
\vspace{-1.5mm}

Here $\exp (\cdot)$ is applied element-wise, $\mathbf{1}_L$ is the all-one vector of length $L$, and $\mathrm{diag} (\cdot)$ is a diagonal matrix with the input vector as the diagonal. The time complexity of computing Eq. (\ref{eq:attnorm1}) is $O(L^2 d)$.

\underline{\textit{Discussions.}} Definition \ref{gen_graph_attention} is highly general because one can flexibly choose the representation the positions $\mathbf{r}_i$ and the function $f$. For example, for sequential data like texts, positional indices in the sequence serves as the positional features ($\mathbf{r}_i=i$), and the RPE mask is a learnable Topelitz matrix ($f(i-j)=c_{i-j}$ with parameters $\{c_{k}\}_{k=-(L-1)}^{L-1}$) \cite{raffel}. For geometric data like 3D molecular structures, one can view $\mathbf{r}_{j}$ as the 3D coordinates $\mathbf{r}_{j}$ of tokens (e.g., atoms) and use some domain-specific $f$ \cite{shi2022benchmarking}. We emphasize that the general formulation is novel and important. It motivates the highly general \FLTs~applicable to a wide range of data and tasks, as opposed to existing approaches that heavily rely on the structure of sequential data and Toeplitz RPE masks (Sec. \ref{sec:related}). 

\paragraph{Kernelized linear attention.} Kernelized attention techniques, e.g., Performers, leverage a decomposition of the attention matrix $\mathbf{A}$ to avoid explicit materialization of $\mathbf{A}$, hence avoid the quadratic complexity in $L$. For the softmax attention, this is achieved by linearizing the softmax kernel $\exp(\mathbf{x}^{\top}\mathbf{y})$ via random features, i.e., constructing for certain randomized mappings $\phi: \mathbb{R}^{d} \rightarrow \mathbb{R}^{m}$ such that $\exp(\mathbf{x}^{\top}\mathbf{y}) = \mathbb{E}[\phi(\mathbf{x})^{\top}\phi(\mathbf{y})]$.

Define $\mathbf{Q}',\mathbf{K}' \in \mathbb{R}^{L \times m}$ as matrices of rows given as $\phi(\mathbf{q}_{i}^{\top}d^{-\frac{1}{4}})^{\top}$ and $\phi(\mathbf{k}_{i}^{\top}d^{-\frac{1}{4}})^{\top}$ respectively. Then the above linearization of softmax kernel directly leads to the following approximate algorithm for attention \textit{without} RPE masks:
\begin{align}
    \widehat{\mathrm{Att}_\mathrm{K}} (\mathbf{Q}, \mathbf{K}, \mathbf{V})&  = \widehat{\mathbf{D}}^{-1} (\mathbf{Q}'(\mathbf{K}'^{\top} \mathbf{V})) \nonumber \\
    \text{where }\widehat{\mathbf{D}} &= \mathrm{diag} (\mathbf{Q}'(\mathbf{K}'^{\top} \mathbf{1}_L) ). \label{performers_attention}
\end{align}    
Here $\widehat{\mathrm{Att}_{\mathrm{K}}}$ stands for the approximate attention and brackets indicate the order of computations. The time and space complexity of this mechanism are $O(Lmd)$ and $O(Lm+md+Ld)$ respectively, compared to $O(L^{2}d)$ and $O(L^{2}+Ld)$ for regular attention. Thus, for $m \ll L $, Performers provide substantial computational improvements.

\section{METHOD}
\label{sec:algorithms}
\subsection{Efficient RPE-enhanced attention}
\label{subsec:flt}
The algorithm presented in Eq. (\ref{performers_attention}) does not incorporate RPE mechanisms. In this subsection, we first present in Theorem \ref{thm:rpe-decompose} a novel technique to derive the (approximate) low rank decomposition of general RPE mask $\mathbf{N}$ in Definition \ref{gen_graph_attention}. Next, we introduce \textrm{FourierLearner-Transformer} (\FLT), a Performer-friendly RPE attention mechanism based on the decomposition. 

\begin{algorithm*}
\caption{FourierLearner Transformer: linear-complexity RPE-enhanced attention}
\label{alg:flt}
\begin{algorithmic}[1] 
\REQUIRE Input queries, keys, values $\mathbf{Q}, \mathbf{K}, \mathbf{V} \in \mathbb{R}^{L \times d}$ and positions $\mathbf{R}\in\mathbb{R}^{L\times \ell}$; random feature map for attention $\phi$ (see Sec. \ref{sec:preliminaries}); Fourier Transform of the RPE function $g_{\theta}$ (potentially parametrized by $\theta$).
\ENSURE Approximate RPE-enhanced attention (Definition \ref{gen_graph_attention})
\STATE \texttt{\# Apply random feature maps for RPE approximation.}
\STATE \texttt{\# $\varphi$ and $\psi$ are defined in Theorem \ref{thm:rpe-decompose} and applied column-wise; $g_{\theta}$ is called in $ \varphi$ and $\psi$.}
\STATE $\mathbf{N}_{1}\leftarrow \varphi(\mathbf{R})$, $\mathbf{N}_{2}\leftarrow \psi(\mathbf{R})$ 
\STATE \texttt{\# Concatenate along the second axis.}
\STATE $\widehat{\mathbf{Q}} \leftarrow [\mathbf{N}_{1},\mathbf{Q}d^{-\frac{1}{4}}] $,
$\widehat{\mathbf{K}} \leftarrow [\mathbf{N}_{2},\mathbf{K}d^{-\frac{1}{4}}]$ 
\STATE \texttt{\# Apply random feature map $\phi$.}
\STATE $\mathbf{Q}' \leftarrow \phi(\widehat{\mathbf{Q}})$, $\mathbf{K}' \leftarrow \phi(\widehat{\mathbf{K}})$
\STATE \texttt{\# Kernelized linear attention (Sec. \ref{sec:preliminaries}). Brackets indicate the order of computations.}
\STATE $\mathbf{B}_1\leftarrow \mathbf{Q}'(\mathbf{K}'^{\top} \mathbf{V})$, $\mathbf{B}_2\leftarrow \mathbf{Q}'(\mathbf{K}'^{\top} \mathbf{1}_L)$, $\mathbf{O}\leftarrow\mathrm{diag}(\mathbf{B}_2)^{-1}\mathbf{B}_1$.
\RETURN $\mathbf{O}$
\end{algorithmic}
\end{algorithm*}

\begin{theorem}
\label{thm:rpe-decompose}
    Given $f:\mathbb{R}^{\ell} \rightarrow \mathbb{R}$ and
    $\mathbf{N} = [f(\mathbf{r}_{i}-\mathbf{r}_{j})] \in \mathbb{R}^{L \times L}$ as defined in Definition \ref{gen_graph_attention}, denote by $g$ the Fourier Transform of $f$. Assume $p$ is some probability density function supported over $\mathbb{R}^{\ell}$. Sample $\xi_1, \cdots, \xi_r\overset{\mathrm{iid}}{\sim} p$ and define the following random feature maps (where $\mathbf{i}=\sqrt{-1}$):
    \begin{align*}
        \varphi(\mathbf{z})=&\tfrac{1}{\sqrt{r}}\left[\mathrm{e}^{2\pi \mathbf{i}\mathbf{z}^{\top}\boldsymbol{\xi}_1}\sqrt{\tfrac{g(\boldsymbol{\xi}_1)}{p(\boldsymbol{\xi}_1)}}, \cdots, \mathrm{e}^{2\pi \mathbf{i}\mathbf{z}^{\top}\boldsymbol{\xi}_r}\sqrt{\tfrac{g(\boldsymbol{\xi}_r)}{p(\boldsymbol{\xi}_r)}}\right]^{\top};\\
        \psi(\mathbf{z})=&\tfrac{1}{\sqrt{r}}\left[\mathrm{e}^{-2\pi \mathbf{i}\mathbf{z}^{\top}\boldsymbol{\xi}_1}\sqrt{\tfrac{g(\boldsymbol{\xi}_1)}{p(\boldsymbol{\xi}_1)}}, \cdots, \mathrm{e}^{-2\pi \mathbf{i}\mathbf{z}^{\top}\boldsymbol{\xi}_r}\sqrt{\tfrac{g(\boldsymbol{\xi}_r)}{p(\boldsymbol{\xi}_r)}}\right]^{\top},
    \end{align*}
    Define $\mathbf{N}_{1}=  \left[\varphi(\mathbf{r}_{1}),\cdots,\varphi(\mathbf{r}_{L})\right]^{\top} \in \mathbb{R}^{L \times r}$ and $\mathbf{N}_{2}=\left[\psi(\mathbf{r}_{1}),\cdots,\psi(\mathbf{r}_{L})\right]^{\top} \in \mathbb{R}^{L \times r}$. Then $\mathbb E [\mathbf{N}_{1}\mathbf{N}_{2}] = \mathbf{N}$.
\end{theorem}

\paragraph{Performer-friendly RPE attention.}Theorem \ref{thm:rpe-decompose} implies that $\widehat{\mathbf{N}}=\mathbf{N}_{1}\mathbf{N}_{2}$ is a low-rank unbiased estimator of $\mathbf{N}$. Consequently, a Performer-friendly RPE attention mechanism with linear complexity can be obtained. Specifically, let $\widehat{\mathbf{Q}} = [\mathbf{N}_{1},\mathbf{Q}d^{-\frac{1}{4}}] \in \mathbb{R}^{L \times (m+r)}$,
$\widehat{\mathbf{K}} = [\mathbf{N}_{2},\mathbf{K}d^{-\frac{1}{4}}] \in \mathbb{R}^{L \times (m+r)}$ where the concatenation is conducted along the second axis. Then
\begin{equation}
\label{eq:Performer-friendly-RPE}
\widehat{\mathbf{A}} \overset{\mathrm{def}}{=} \exp\left(\widehat{\mathbf{N}}+\frac{\mathbf{Q}\mathbf{K}^{\top}}{\sqrt{d}}\right)
= \exp\left(\widehat{\mathbf{Q}}\widehat{\mathbf{K}}^{\top}\right).
\end{equation}
In Eq. (\ref{eq:Performer-friendly-RPE}), RPE-masked attention is now translated to regular softmax attention that admits ``Performerization'' as described in Eq. (\ref{performers_attention}). This observation naturally leads to an efficient RPE-enhanced attention algorithm, with a pseudo-code implementation provided in Algorithm \ref{alg:flt}. The time and space complexity of the algorithm are $O(L(m+r)d)$ and $O(L(m+r)+(m+r)d+Ld)$, respectively.

In Algorithm \ref{alg:flt}, instead of learning $f$ and trying to compute its Fourier Transform $g$ for the low-rank decomposition of $\mathbf{N}$, we propose to directly learns $g$ and refer to our approach as \textbf{FourierLearner-Transformer} (\textbf{FLT}). Note that \FLT~effectively learns a spectral representation of $f$. 

We point out that our formulations are general enough to cover a wide range of RPE variants used in practice:

\alertinline{Regular RPE for sequential data.}
In this setting the input sequence does not have richer geometric structure and thus vectors $\mathbf{r}_{j}$ can be identified as the indices of tokens in the sequence, i.e., $\mathbf{r}_{j}=j$. Thus, \FLT~learns a function $g:\mathbb{R} \rightarrow \mathbb{C}$ (Sec. \ref{sec:lm}, \ref{sec:images}).

\alertinline{RPE for 3D-data.} For this input type (e.g. 3D molecular data), it is natural to identify $\mathbf{r}_{j}$ as the 3D coordinates of atoms. Thus, \FLT~learns a function $g:\mathbb{R}^{3} \rightarrow \mathbb{C}$. Note that existing methods \cite{liutkus, rpe-performers, topmasking} are \textit{inapplicable} while \FLT~works well in this setting (Sec. \ref{sec:molecular_dynamics}).

Finally, we note that \FLT~necessitates specifying some distribution $p$ supported over $\mathbb{R}^{\ell}$ to satisfy the assumption in Theorem \ref{thm:rpe-decompose}.  Practical considerations dictate that $p$ needs to be chosen in such a way that we can efficiently sample from it and compute its density function. In our experiments, we use Gaussian distributions zero mean and unit variance/learnable variance for $p$.

\subsection{Theoretical analysis of FLT}
We have theoretically investigated \FLT 's RPE approximation. In particular, we prove the following theorem, which states that under mild assumptions, the estimated RPE mask $\widehat{\mathbf{N}}$ can approximate the true RPE mask $\mathbf{N}$ up to arbitrary precision with high probability. Besides, the theorem provides sample complexity bound for such accurate approximation.

\begin{theorem}[Uniform convergence and sample complexity for approximation]\label{thm:sample-complexity-flt}
    Given $L$ vectors $\mathbf{r}_{1},...,\mathbf{r}_{L} \in \mathbb{R}^{\ell}$, define the RPE attention mask $\mathbf{N} = [f(\mathbf{r}_{i}-\mathbf{r}_{j})]_{i,j\in [L]}$.   
    Assume that $c= \||g(\mathbf{x})|/p(\mathbf{x})\|_{\infty}$, where $g$ is the Fourier Transform of $f$ and $p$ is some probability density function over $\mathbb{R}^{\ell}$.
    
    For any $\varepsilon, \delta>0$, if the number of random features $r=\Theta \left(\frac{c^2}{\varepsilon^2}\log\frac{L}{\delta}\right)$, then \FLT 's RPE approximator $\widehat{\mathbf{N}}$ satisfies 
    $$\mathbb{P}\left(\|\mathbf{N}-\widehat{\mathbf{N}}\|_{\max}\leq \varepsilon \right)>1-\delta,$$
    where $\|\cdot\|_{\max}$ denotes the max norm of a matrix.
\end{theorem}

We also prove variance bound of the estimated RPE and present the result in the supplementary material. The proofs and detailed discussions on the theoretical results can be found in the supplementary material as well.

\subsection{The topology of the Fourier Transform}
\label{sec:topology}

Nowhere in the analysis in Sec. \ref{subsec:flt} have we relied on any structural properties of $f$. In particular, the matrix $\mathbf{N}$ does not need to be a valid positive definite kernel-matrix or even symmetric. However, if needed, desired inductive bias can be incorporated into \FLT~via certain parameterization schemes used to train $g$, as we discuss in this subsection.

\paragraph{Gaussian mixture RPEs.} One of the most general parameterizations of $g$ that we have considered is the so-called \textit{Gaussian mixture} variant:
\begin{equation*}
g(\boldsymbol{\xi}) = \sum_{t=1}^{T} w_{t} \exp\left(-\frac{\|\boldsymbol{\xi}-\boldsymbol{\mu}_{t}\|^{2}}{2 \sigma_{t}^{2}}\right).\label{eq:gaussian-mixture-rpe}
\end{equation*}
Therefore, the FT $g$ is parameterized by $(2+\ell)T$ numbers: $w_{1},...,w_{T},\sigma_{1},...,\sigma_{T} \in \mathbb{R}$, $\boldsymbol{\mu}_{1},...,\boldsymbol{\mu}_{T} \in \mathbb{R}^{\ell}$. In the special case where $T=1$, the FT becomes a renormalized Gaussian kernel and as such, defines $f$ as another Gaussian kernel. 

\paragraph{Shift-invariant kernels for RPE masks.} It is straightforward to apply the \FLT~mechanism for RPEs to make $\mathbf{N}$ a kernel-matrix of any \textit{shift-invariant} kernel \cite{rahimi}. By Bochner's Theorem, for a shift-invariant kernel: $\mathrm{K}: \mathbb{R}^{\ell} \times \mathbb{R}^{\ell}  \rightarrow \mathbb{R}$, there exists a corresponding probabilistic distribution $p_{\mathrm{K}}$ and some positive constant $C>0$, such that
\begin{equation*}
\mathrm{K}(\mathbf{x},\mathbf{y}) = C\int_{\mathbb{R}^{d}} e^{\mathbf{i}(\mathbf{x}-\mathbf{y})^{\top}\boldsymbol{\xi}}p_{\mathrm{K}}(\boldsymbol{\xi})\mathrm{d}\boldsymbol{\xi}
\end{equation*}
Thus, to obtain an unbiased approximation of the RPE mask $\mathbf{N}$ given by the kernel matrix $[\mathrm{K}(\mathbf{s}_{i},\mathbf{s}_{k})]_{i,k=1,...,L}$ for the shift-invariant kernel $\mathrm{K}$, it suffices to take $\mathbf{r}_{j}=\frac{1}{2\pi}\mathbf{s}_{j}$,  
$g(\boldsymbol{\xi})=Cp_{\mathrm{K}}(\boldsymbol{\xi})$ for $j=1,....r$. Even if a particular class of
shift-invariant kernels has been chosen, \FLT~still provides a way to learn its specific instantiation through learning an appropriately parameterized $g$. 

\paragraph{Local RPEs.} Through the corresponding structured parameterized Fourier Transforms $g$, \FLT~is also capable of modeling various schemes where the RPE mechanism needs to be applied only locally and regular attention is to be used for tokens far enough from each other. We call such strategies \textit{local RPEs}. Local RPEs can be derived for both sequential data and high-dimensional geometric data.

The most basic local RPE takes $\mathbf{r}_{j}=j$ and, for an attention radius $v>0$ and $C \in \mathbb{R}$, defines $f$ as\footnote{Note that instead of using one indicator function in Eq. (\ref{eq:sin}), one can also apply a linear combination of many with learnable radii and a list of coefficients.}
\begin{equation}
\label{eq:sin}
f_{v, C}(\Delta r) = C \cdot \mathbb{I} [|\Delta r| \leq v].
\end{equation}
Such an RPE mechanism would (de)amplify the regular attention score between tokens close to each other by a certain multiplicative amount and might play a similar role as local attention \cite{local_attention}. It turns out that the FT for such a $f$ has a particularly simple form:
$$
g_{f_{v, C}}(\xi) = C \cdot \frac{\sin(2\pi v\xi)}{\pi \xi}.
$$

Interestingly, RPEs from Eq. (\ref{eq:sin}) can be easily generalized to a higher-dimensional local RPE. In this case, we consider the positional encoding function $f:\mathbb{R}^{\ell}\to \mathbb{R}$ of the following form:
\begin{equation*}
f_{\mathbf{v}, C}(\Delta \mathbf{r}) =  \prod_{j=1}^{\ell} C \cdot \mathbb{I}[|\Delta r^{(j)}| \leq v_{j}]\quad (\Delta \mathbf{r}\in\mathbb{R}^{\ell}),
\end{equation*}

where $\Delta r^{(j)}$ denotes the $j$-th entry of $\Delta \mathbf{r}$. The corresponding Fourier Transform $g$ can be factorized as 
$$
g_{f_{\mathbf{v}, C}}(\boldsymbol{\xi}) = C \cdot \prod_{j=1}^{\ell} \frac{\sin(2\pi v_{j}\xi_{j})}{\pi \xi_{j}}.
$$
This result can be further generalized. Consider the function $g$ of the following form: 
\begin{equation*}
g_{k_{1},...,k_{\ell}}^{v_{1},...,v_{\ell}}(\boldsymbol{\xi}) = C \cdot \prod_{j=1}^{\ell} \frac{\sin^{k_{j}}(2\pi v_{j}\xi_{j})}{\pi \xi_{j}}
\end{equation*}
The inverse Fourier Transform of $g$ can be written as
$$
    f(\Delta \mathbf{r}) = M \cdot \prod_{j=1}^{d}f_{j}^{v_{j}}(\Delta r_j),
$$
where $M$ is a constant and each $f^{v_{j}}_{j}$ is (a) continuous, (b) symmetric, (c) with compact support of length depending on $v_{j}$, and (d) piece-wise a polynomial of order $k_{j}-1$.
Such functions $f$ are natural candidates for continuous local RPE mechanisms for tokens with positions embedded in $\mathbb{R}^{\ell}$ and any $\ell \geq 1$. 
Examples of local RPE variants for $\ell=2$, supported via \FLT, are presented in Fig. \ref{fig:local} in Appendix \ref{sec:visual-local}.

The above theoretical results can be directly obtained via straightforward integration and a realization that the $N$-dim FT of a function: $h(x_{1},\cdots,x_{N}) \overset{\mathrm{def}}{=} h_{1}(x_{1})\cdot\cdots\cdot h_{N}(x_{N})$ can be represented as the product of 1D FTs of the individual components $h_{j}$.

\paragraph{Remark.} We point out that all the three parametrization schemes above are parameter-efficient. In all our experiments, \FLT~introduced $<0.03$M additional parameters for relative positional encoding. Note that the number of additional parameters does not increase with the input sequence length.

\section{EXPERIMENTS}
\label{sec:expeeriments}

In this section, we provide experimental results on diverse tasks to demonstrate the effectiveness of the \FLT~architecture. We first study language modeling with sequential text data, which is a standard setting for efficient RPE-enhanced attention and enables thorough comparisons with existing baselines. Next, we consider the computer vision domain and test \FLTs~on several image classification datasets. Finally, to show that \FLTs~ broaden the scope of RPE-enhanced efficient Transformers, we experiment on molecular property prediction with complicated RPE masks that existing efficient RPE-enhanced attention baselines cannot handle. The complete experimental setup, the hyper-parameters for each of the tasks, and hardware details are provided in the supplementary material.

\begin{table}
    \begin{center}
    \caption{\textbf{Language model perplexity scores} on the WikiText-103 validation set. The lowest perplexity is highlighted in \textbf{bold}.}\label{tab:lm}
    \scalebox{0.95}{\begin{tabular}{@{}lc@{}}
    \toprule
    Model &  Perplexity \\ \midrule
    Linear Trans. \cite{katharopoulos2020transformers}  &38.4\\
    RFA-Gaussian \cite{peng2021random}&  33.6 \\
    RFA-arccos \cite{peng2021random}& 36.0 \\
    RFA-GATE-Gaussian \cite{peng2021random}& 31.3 \\
    RFA-GATE-arccos \cite{peng2021random} & 32.8 \\
    Performer \cite{choromanski}& 31.1\\
    CosFormer \cite{cosformer}& 30.7\\\midrule
    Performer-sineSPE \cite{liutkus} & 38.0 \\
    Performer-convSPE \cite{liutkus} & 37.8 \\
    Log-linear Performer \cite{rpe-performers}  &30.6\\\midrule
    \FLT~(Gaussian mixture RPE) (ours) & 30.3 \\
    \FLT~(local RPE) (ours) & \textbf{30.1}\\ \bottomrule
    \end{tabular}
    }
    \end{center}
\end{table}

\subsection{Language modeling}
\label{sec:lm}

We conduct experiments on the WikiText-103 language modeling task to show the effectiveness of our proposed method in NLP applications. Most existing baselines are applicale to sequential text data. Thus, we provide comprehensive empirical comparisons on model quality and efficiency with baselines in this subsection.

\paragraph{Compared methods.}
In this experiment, we study \FLT~with two RPE vaiants, Gaussian mixture RPE and local RPE. We compare our model with the following strong baselines:

\begin{itemize} 
\item \textit{Linear Transformer} \cite{katharopoulos2020transformers}, which uses kernelized low-rank attention with $\mathrm{elu}(\cdot)+1$ as the feature map. 
\item  \textit{Random feature attention} (RFA) {\small \cite{peng2021random}}, which has two variants (Gaussian and arc-cosine) and an optional gating mechanism. 
\item  The regular \textit{Performer} \cite{choromanski}, which applies the FAVOR+ mechanism for attention matrix approximation. 
\item  \textit{CosFormer} \cite{cosformer}, which designs a linear operator and a cosine-based distance re-weighting mechanism for attention matrix approximation. 
\item \textit{Performer-SPE} \cite{liutkus}, which incorporates a special class of RPE into low-rank attention and has two variants (sineSPE and convSPE). 
\item  The \textit{log-linear Performer} \cite{rpe-performers} which extends Performers to work with an arbitrary Toeplitz RPE attention mask.
\end{itemize}

\begin{figure*}[!t]
    \centering
    \includegraphics[width=0.48\linewidth]{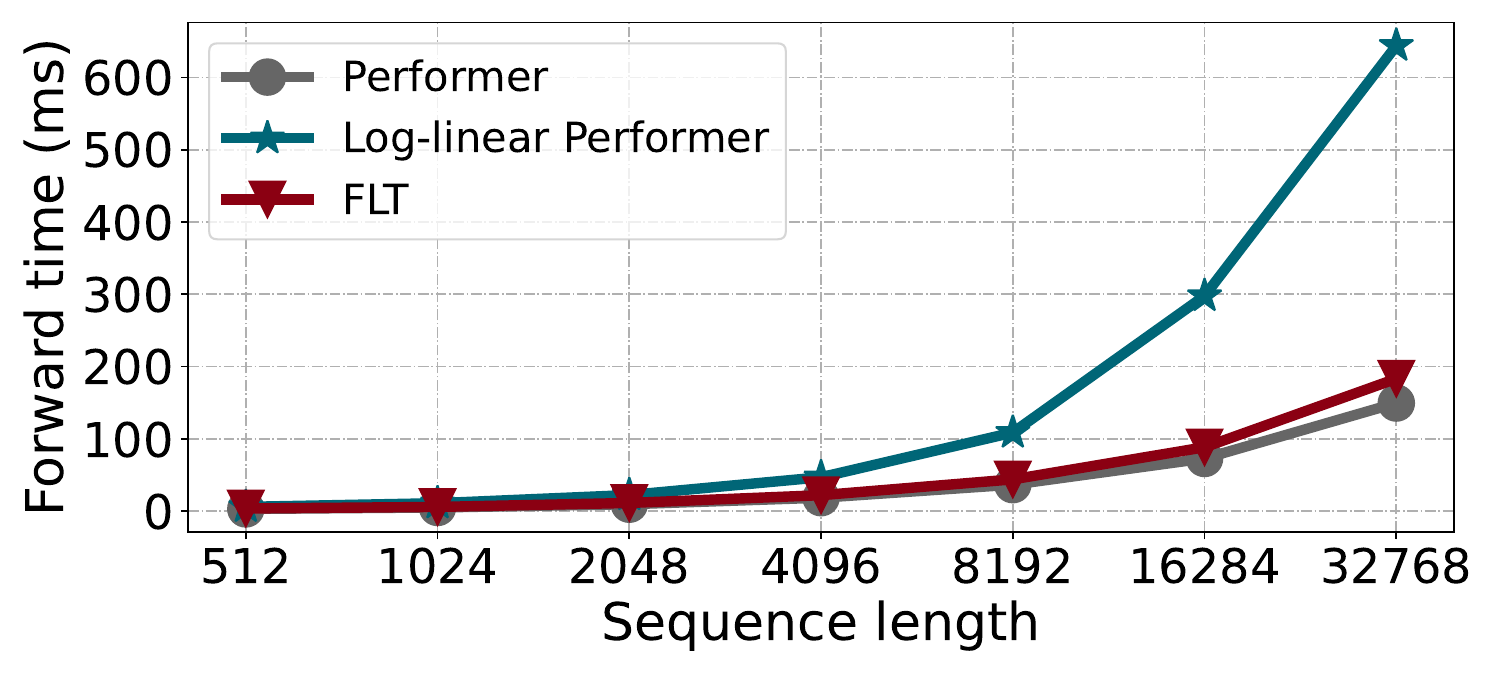}
    \includegraphics[width=0.48\linewidth]{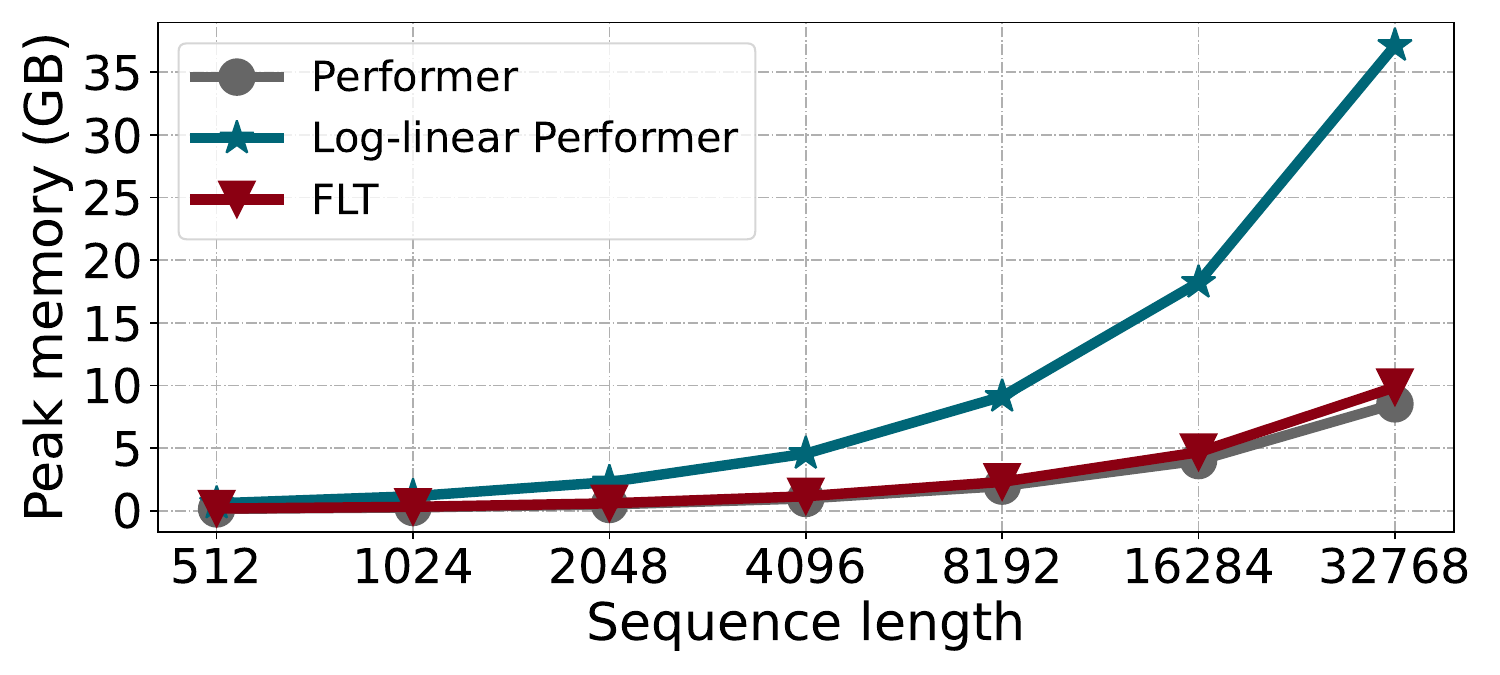}
    \vspace{-4mm}  
    \caption{\textbf{Model forward speed} (left) \textbf{and peak memory} (right) \textbf{comparisons} between \FLT~and baselines under different input sequence lengths.}
    \vspace{-3mm}
    \label{fig:efficiency}
\end{figure*}

\paragraph{Implementation details.} All the tested models are efficient Transformers based on kernelized low-rank attention, with 6 decoder layers. More details regarding model configurations and training are in the supplementary material. We use the validation perplexity as the evaluation metric; lower perplexity indicates better performance.

\paragraph{Results.}
The results are shown in Table \ref{tab:lm}. Both variants of our \FLT~outperform all the baselines. Compared with efficient Transformers without RPE, \FLT~achieves much stronger performance. For example, the validation perplexity of our \FLT~with local RPE is 1.0 point lower than that of the regular Performer, indicating that our method brings substantial performance gains by incorporating RPE into the attention module.

Compared with other efficient Transformer variants \textit{with} RPE, our \FLT~is still very competitive. For example, our \FLT~achieves lower perplexity than the strong log-linear Performer baseline. Note that log-linear Performer relies on more expensive FFT and is less efficient in practice. Specifically, the time and space complexity of the \FLT~are $O(L(m+r)d)$ and $O(L(m+r)+(m+r)d+Ld)$, respectively, while the time and space complexity of log-linear Performer are $O(Lmd\log L)$ and $O(Lmd)$.
Thus, our \FLT~obtains both better quality and efficiency than existing efficient RPE-enhanced Transformer variants on this task. 

In addition, we further investigate the attention matrices of \FLT\footnote{\FLT~does not explicitly construct attention matrices during training so that it avoids the quadratic computational complexity. However, we can still materialize the attention matrices approximated by \FLT~}. We visualize the attention matrices of different attention heads in an \FLT~model trained on WikiText-103 language modeling in Fig. \ref{fig:attn-mat} in Appendix \ref{sec:attn-mat}. The visualizations show that some attention heads pay more attention to nearby tokens, while others shows global attention patterns. Quantitatively, the average attention probability over the most distant/nearby 10\% tokens is 0.068/0.279 respectively. Thus, \FLT~ learns locality bias in language while maintaining the advantage to capture global contexts and leverage information in distant tokens.

\paragraph{Computational cost comparisons.} 
As discussed above, \FLT~enjoys much better time/space complexity compared with the strongest baseline method, the log-linear Performer. To showcase \FLT 's efficiency in practice, we construct one Transformer layer with 12 attention heads whose hidden dimension is set to 768, and FFN dimension is set to 3072. 
We feed inputs with varying lengths and a batch size of 8 into the model and measure the efficiency. We report the average forward time and the maximum peak memory consumption across 5 runs under different input sequence lengths in Fig. \ref{fig:efficiency}. 
We compare \FLT~with the strongest baseline, log-linear Performer, and we also include the regular Performer as a reference.
It's clear that \FLT~only introduces negligible memory overhead compared with the regular Performer, and scales much better than the log-linear Performer in practice, in terms of both model forward time and peak memory. Therefore, our experiment show that \FLT~is both more accurate and more scalable than the baselines on sequential text data.

\begin{table}[t]
  \centering
  \caption{\textbf{Image classification accuracy} comparisons. Log-linear Performer is omitted due to its infeasible memory complexity and out-of-memory issues. The best performances are highlighted in \textbf{bold}.}
  \label{tab:image}
  \begin{tabular}{@{}lccc@{}}
    \toprule
    & ImageNet & Places365 & FashionMnist \\
    \midrule
    Performer & 75.1\% & 55.0\% & 91.1\% \\
    CosFormer & 76.2\% & 55.6\% & 91.6\% \\
    FLT (ours) & \textbf{77.4}\% & \textbf{56.0}\% & \textbf{92.1}\% \\
    \bottomrule
  \end{tabular}
\end{table}

\begin{figure*}[!ht]
\begin{minipage}[t]{0.51\linewidth}
    \centering
    \vspace{-43mm}
    \captionof{table}{\textbf{Comparisons of FLT with the regular Performer} on OC20 IS2RE task. The suffix ``-$k$L'' means the model consists of $k$ layers, e.g., \FLT -10L refers to a 10-layer \FLT. The evaluation metrics are Mean Absolute Error (MAE, lower is better) of the energies and the percentage of Energies within a Threshold (EwT, higher is better). We highlighted in \textbf{bold} the best performance.}\label{tab:oc20}
    \vspace{-1.5mm}
    \scalebox{0.90}{
        \begin{tabular}{lcc} \toprule
                        & Energy MAE (eV) & EwT (\%) \\ \midrule
        Performer-12L       & 0.5454          & 4.90      \\
        \FLT -10L (ours)  & 0.5157          & \textbf{5.44}     \\
        \FLT -12L (ours)  & \textbf{0.5046}          & 5.33    \\ \bottomrule
        \end{tabular}
    }
\end{minipage}\hfill
\begin{minipage}[t]{0.46\linewidth}
    \centering
    \includegraphics[width=0.99\linewidth]{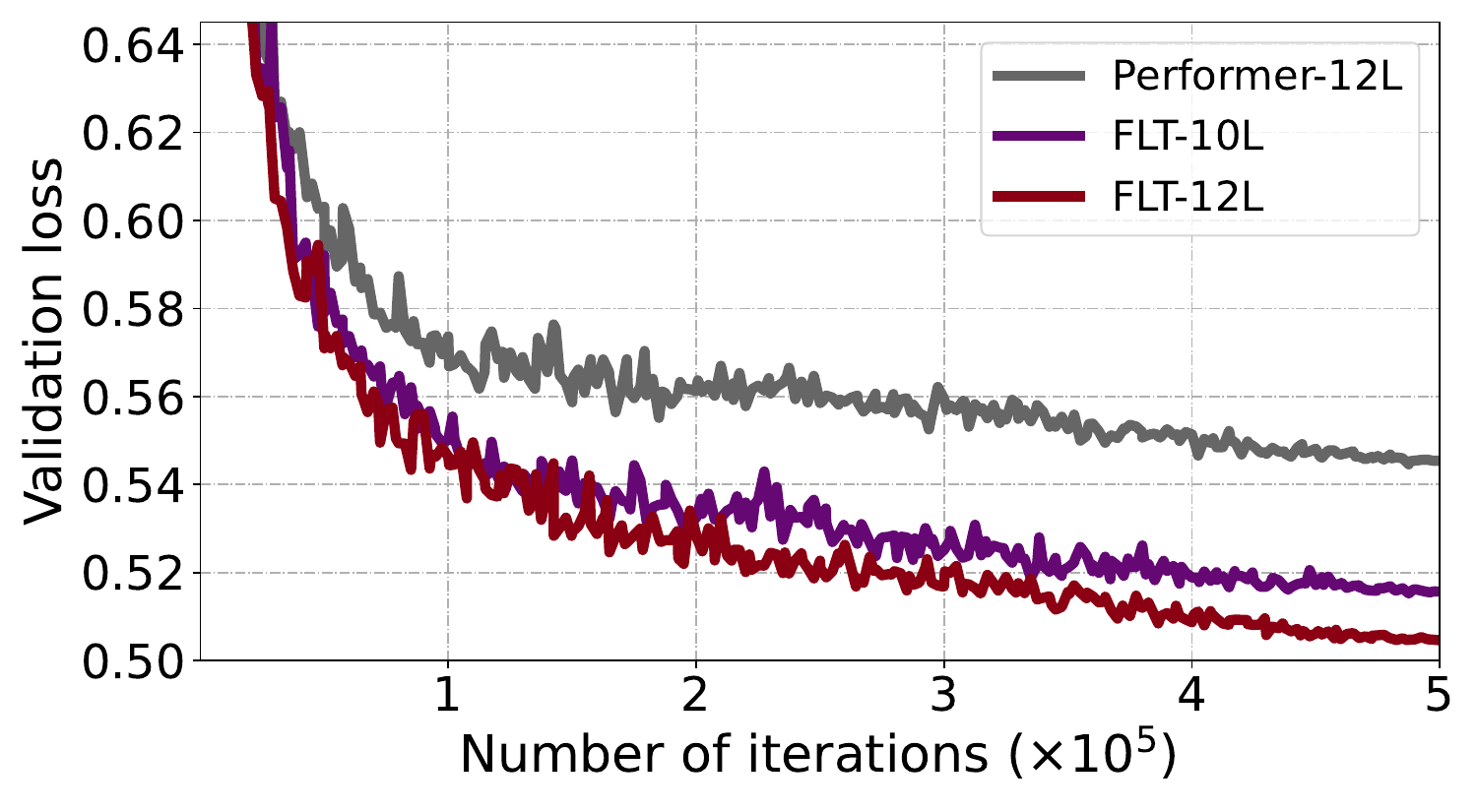}
    \vspace{-7.5mm}  
    \caption{\textbf{Validation loss} of FLTs and the regular Performer on the IS2RE task of OC20 dataset.}
    \label{fig:oc20}\vspace{-6mm}
\end{minipage}
\end{figure*}

\subsection{Image classification}
\label{sec:images}
We thoroughly benchmarked \FLT~variants of Vision Transformers (ViTs) \cite{vits} on several image classification datasets, including ImageNet, Places365, and FashionMnist. Details of these datasets can be found in the supplementary material.

\paragraph{Compared methods and implementation details.} We compare \FLT~with the regular Performer as well as the most competitive competitor from Sec. \ref{sec:lm}, \textit{CosFormer} and \textit{log-linear Performer}. All tested ViTs consist of 12 layers with 12 attention heads in each layer. More details regarding model configurations and training are in the supplementary material.
For our \FLT~variants, we apply Gaussian mixture RPEs (Sec. \ref{sec:topology}) with the number of Gaussian mixture modes $T$ set to 25 and the number of random features for RPE-encoding $r$ set to 64. 

\paragraph{Results.}
The results are presented in Table \ref{tab:image}. The \textit{log-linear Performer} architecture run out of memory for $m=128$ and does not train when $m$ was reduced (with a fixed batch size of 4096) to fit the assigned memory. Thus, it is omitted in the comparison. Compared with the other two baselines, our \FLT~obtains strongest performances on all the three datasets. For instance, on ImageNet, \FLT~provides a $\mathbf{2.3}$\% accuracy improvement over the regular Performer; and is even $\mathbf{1.2}$\% better than the strong CosFormer baseline. The results demonstrate that \FLT~also works well on image data.




\subsection{Molecular property prediction}
\label{sec:molecular_dynamics}

As highlighted in previous discussions, \FLT~broadens the scope of RPE-enhanced efficient Transformers and can be applied to geometric data embedded in high-dimensional Euclidean spaces. To validate this claim, in this subsection, we further evaluate our \FLT~model on the molecular property prediction task to show its capability to handle 3D input data and complicated (non-Toeplitz) RPE masks. To the best of our knowledge, in this scenario, \FLT~is the \textit{first} Transformer providing RPE-enhanced scalable attention that enjoys \textit{linear} complexity with respect to the number of input tokens.

We use a publicly-available large-scale electrocatalysts dataset - the Open Catalyst 2020 (OC20) dataset and focus on the IS2RE task which requires to predict the energy of the relaxed structure given the initial structure of solid catalysts with adsorbate molecules \cite{ocp_dataset}.

\paragraph{Compared methods.}
Existing technicques considered in the previous experiments do not apply to this setting. Thus, we only compare our \FLT~with the regular Performer without RPE. For the \FLT~model, we consider to approximate RPE masks based on Gaussian basis functions, which are popularly used in neural networks for molecular modeling \cite{gasteiger2021gemnet, shi2022benchmarking, luo2023one}. Specifically, the RPE mask is defined as $\mathbf{N} = [f(\mathbf{r}_{i}-\mathbf{r}_{j})]_{i,j\in[L]} \in \mathbb{R}^{L \times L}$, where $\mathbf{r}_{i}\in \mathbb{R}^{3}$ is the position of the $i$-th input atom, $L$ is the total number of input atom, and
$$
    f(\mathbf{r}) = \sum_{t=1}^T \frac{w_t}{(\sqrt{2\pi}\sigma_t)^3} \exp\left(-\frac{\|\mathbf{r}\|^2}{2\sigma_t^2}\right).
$$
Note that  RPE only calculates the relative distances between atoms, which naturally preserves many invariant and equivariant properties.
It easy to see that the Fourier Transform of $f$ is 
$$
    g (\boldsymbol{\xi})=\sum_{t=1}^T w_t\exp\left(-2\pi^2\sigma_t^2\|\boldsymbol{\xi}\|^2\right),
$$
which enables us to approximate the RPE mask $\mathbf{N}$ in \FLTs~using the technique described in Sec. \ref{sec:algorithms}.

\begin{figure*}
    \centering
    \includegraphics[width=0.95\linewidth]{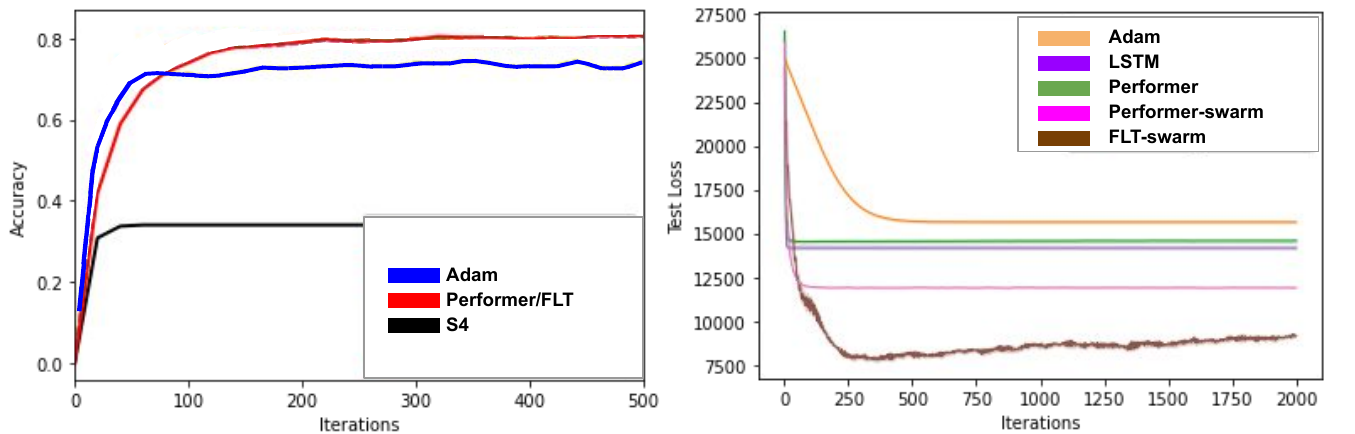}
    \caption{\textbf{Results of learnable optimizer experiments. } \textbf{Left:} \textrm{Adam} \& learnable optimizers using \FLT~and S4 on the task of training ViT-Base classifier on ImageNet. \textbf{Right:} Adam \& various learnable optimizers on the task of optimizing Rastrigin-type functions (from private conversation with the authors of \cite{jain2023mnemosyne}).}
    \label{fig:learn-opt}
\end{figure*}

\paragraph{Implementation details.} We adopt most of the training strategies of 3D-Graphormer \cite{shi2022benchmarking}. 
Specifically, we trained a regular Performer with 12 layers and two \FLT~with 10 and 12 layers respectively. 
More details regarding model configurations and training are in the supplementary material.
We evaluate the model performance on the in-domain validation set. We use Mean Absolute Error (MAE) of the energies and the percentage of Energies within a Threshold (EwT) of the ground truth energy to evaluate the accuracy of the predicted energies. 

\paragraph{Results.}
The results are presented in Table \ref{tab:oc20}. We also present the validation loss curves of the models in Fig. \ref{fig:oc20} for a more comprehensive comparison. 
Clearly, our \FLT~models obtain better performance in both evaluation metrics and produce more accurate energy predictions. For example, the energy MAE of the 12-layer \FLT~is more than 0.04eV lower than that of the 12-layer regular Performer, which indicates that the use of RPE effectively increases the predictive power of the model. One may argue that the use of RPE in \FLT~may add some computational overhead and increase the number of model parameters. However, it should be noted that a shallower 10-layer \FLT~can also significantly outperform the 12-layer regular Performer, while being faster and using less parameters. 

\subsection{Learnable optimizers}
\label{sec:mnemosyne}

\FLT~has also been compared independently by authors and other researchers on longer contexts with other classes of efficient architecture, including LSTM \cite{hochreiter1997long} and state-space models \cite{s4}. 
The corresponding task is practical and challenging: applying Transformers as memory models in learnable optimizers (with context length up to 2000). 
In this setting, long-range temporal (to understand the \textit{history} of the optimization) and spatial (to understand the \textit{landscape} of the loss function more globally) reasoning is critical \cite{jain2023mnemosyne,gartner2023transformer}. 

\paragraph{Compared methods and implementation details.}
In the first experiment, \FLT-based learnable optimizer is compared against S4-based learnable optimizer \cite{s4} and standard non-learnable Adam optimizer on ImageNet classification.

In the second experiment, \FLT~is further applied on population(swarm)-based learnable optimizers, in which the masking mechanism implemented by FLT was applied to modulate how the members of the population attend to each other. The baselines include standard non-learnable Adam optimizer as well as learnable optimizers based on LSTM and Performer. The evaluation is conducted on Rastrigin-like functions. 

\paragraph{Results.} 
The results of the first/second experiment are shown in the left/right panel of Fig. \ref{fig:learn-opt}. We note that in both experiments, \FLT-based approach provides drastic improvements over all other variants. The results shows the consistent effectiveness of our model in capturing long range dependencies in learnable optimizers.

\section{CONCLUSIONS}
\label{sec:conclusion}
We introduce \textrm{FourierLearner-Transformers} (\FLTs) that efficiently adapt the relative positional encoding (RPE) mechanism into Performers - kernelized implicit-attention Transformers with linear space and time complexity. 
In contrast to other architectures combining Performers with RPEs, \FLTs~maintain linear complexity of the attention modules with no additional structural assumptions regarding the RPE mask. 
We provide theoretical analysis and show that \FLTs~ can accurately approximate RPE. 
We further conduct extensive experiments to show the efficiency and quality of \FLTs~across a wide range of tasks and data modalities, including texts, images, molecules, and optimizer memory.

\section*{ACKNOWLEDGEMENTS}
\label{sec:ac}

We would like to thank Deepali Jain for a discussion on using \FLT~for learnable optimizers, as well as proposing and performing experiments with population(swarm)-based methods for the \FLT~variants provided by the authors. We also thank the reviewers for their helpful comments.

This work is supported in part by the United States Department of Energy via the Brookhaven National Laboratory under Contract No. 384608.

\bibliography{references}

\appendix
\onecolumn
\section{OMITTED THEORETICAL RESULTS AND PROOFS}
\label{app:theory}
We provide here omitted theoretical results, proofs, and discussions on \FLT 's RPE approximation. We first prove the RPE approximation proposed in the main paper is unbiased. Then we provide a variance bound and an approximation guarantee for it.

For convenience of reading, we always state the theorem before providing the proof, even if the theorem has appeared in the main body of the paper.

\subsection{Unbiased RPE approximation}
\begin{theorem}
    \label{appthm:rpe-decompose}
    \footnote{This is Theorem {\color{blue} 4.1} in the main paper.}
    Given $f:\mathbb{R}^{\ell} \rightarrow \mathbb{R}$ and
    $\mathbf{N} = [f(\mathbf{r}_{i}-\mathbf{r}_{j})] \in \mathbb{R}^{L \times L}$ as defined in Definition {\color{blue} 3.1}, denote by $g$ the Fourier Transform of $f$. Assume $p$ is some probability density function supported over $\mathbb{R}^{\ell}$. Sample $\xi_1, \cdots, \xi_r\overset{\mathrm{iid}}{\sim} p$ and define the following random feature maps (where $\mathbf{i}=\sqrt{-1}$):
    \begin{align*}
        \varphi(\mathbf{z})=&\frac{1}{\sqrt{r}}\left[\mathrm{e}^{2\pi \mathbf{i}\mathbf{z}^{\top}\boldsymbol{\xi}_1}\sqrt{\frac{g(\boldsymbol{\xi}_1)}{p(\boldsymbol{\xi}_1)}}, \cdots, \mathrm{e}^{2\pi \mathbf{i}\mathbf{z}^{\top}\boldsymbol{\xi}_r}\sqrt{\frac{g(\boldsymbol{\xi}_r)}{p(\boldsymbol{\xi}_r)}}\right]^{\top};\\
        \psi(\mathbf{z})=&\frac{1}{\sqrt{r}}\left[\mathrm{e}^{-2\pi \mathbf{i}\mathbf{z}^{\top}\boldsymbol{\xi}_1}\sqrt{\frac{g(\boldsymbol{\xi}_1)}{p(\boldsymbol{\xi}_1)}}, \cdots, \mathrm{e}^{-2\pi \mathbf{i}\mathbf{z}^{\top}\boldsymbol{\xi}_r}\sqrt{\frac{g(\boldsymbol{\xi}_r)}{p(\boldsymbol{\xi}_r)}}\right]^{\top},
    \end{align*}
    Define $\mathbf{N}_{1}=  \left[\varphi(\mathbf{r}_{1}),\cdots,\varphi(\mathbf{r}_{L})\right]^{\top} \in \mathbb{R}^{L \times r}$ and $\mathbf{N}_{2}=\left[\psi(\mathbf{r}_{1}),\cdots,\psi(\mathbf{r}_{L})\right]^{\top} \in \mathbb{R}^{L \times r}$. Then 
    \begin{equation*}
        \mathbb E [\mathbf{N}_{1}\mathbf{N}_{2}] = \mathbf{N}.
    \end{equation*}
\end{theorem}

\begin{proof}
    Be definition of $\mathbf{N}$, it suffices to show that
    \begin{equation*}
        f(\mathbf{r}_i-\mathbf{r}_j)=\mathbb{E}\left[\varphi(\mathbf{r}_{i})^{\top}\psi(\mathbf{r}_{j})\right].
    \end{equation*}

    Note that $g$ is the Fourier Transform of $f$. Therefore,
    \begin{align}
        &f(\mathbf{x})=\int_{\mathbb{R}^d}\mathrm{e}^{2\pi\mathrm{i}\mathbf{x}^{\top}\boldsymbol{\xi}}g(\boldsymbol{\xi})\mathrm{d}\boldsymbol{\xi}
        =\int_{\mathbb{R}^d}\mathrm{e}^{2\pi\mathrm{i}\mathbf{x}^{\top}\boldsymbol{\xi}} \frac{g(\boldsymbol{\xi})}{p(\boldsymbol{\xi})}\cdot p(\boldsymbol{\xi}) \mathrm{d}\boldsymbol{\xi}
        =\mathbb{E}_{\boldsymbol{\xi}\sim p} \left[\mathrm{e}^{2\pi\mathrm{i}\mathbf{x}^{\top}\boldsymbol{\xi}}\frac{g(\boldsymbol{\xi})}{p(\boldsymbol{\xi})}\right]. \label{eq:fourier}\\
        \Rightarrow\quad & f(\mathbf{r}_i-\mathbf{r}_j) = \mathbb{E}_{\boldsymbol{\xi}\sim p}\left[\mathrm{e}^{2\pi\mathrm{i}\mathbf{r}_i^{\top}\boldsymbol{\xi}}\sqrt{\frac{g(\boldsymbol{\xi})}{p(\boldsymbol{\xi})}}\cdot \mathrm{e}^{-2\pi\mathrm{i}\mathbf{r}_j^{\top}\boldsymbol{\xi}}\sqrt{\frac{g(\boldsymbol{\xi})}{p(\boldsymbol{\xi})}}\right] .\nonumber
    \end{align}

    In the mean time, by definition of $\varphi$ and $\psi$, we have
    \begin{equation*}
        \varphi(\mathbf{r}_{i})^{\top}\psi(\mathbf{r}_{j}) = \frac{1}{r}\sum_{k=1}^r \mathrm{e}^{2\pi\mathrm{i}\mathbf{r}_i^{\top}\boldsymbol{\xi}_k}\sqrt{\frac{g(\boldsymbol{\xi}_k)}{p(\boldsymbol{\xi}_k)}}\cdot \mathrm{e}^{-2\pi\mathrm{i}\mathbf{r}_j^{\top}\boldsymbol{\xi}_k}\sqrt{\frac{g(\boldsymbol{\xi}_k)}{p(\boldsymbol{\xi}_k)}} 
    \end{equation*}

    Finally, note that $\boldsymbol{\xi}_1, \cdots, \boldsymbol{\xi}_m \sim \mathrm{i.i.d.}~p$. By linearity of expectation, we have $ f(\mathbf{r}_i-\mathbf{r}_j)=\mathbb{E}\left[\varphi(\mathbf{r}_{i})^{\top}\psi(\mathbf{r}_{j})\right]$ and conclude the proof.
\end{proof}


\subsection{Variance of RPE approximation}

\begin{lemma}\label{applemma:var-flt}
    Assume that $c= \||g(\mathbf{x})|/p(\mathbf{x})\|_{\infty}$, where $g$ is the Fourier Transform of the RPE function $f$ and $p$ is the probability density function of some probabilistic distribution. Then the following is true for any $\mathbf{z}\in\mathbb{R}^{\ell}$:
    \begin{equation}\label{eq:var-flt}
        \mathrm{Var}_{\boldsymbol{\xi}\sim p}\left[\mathrm{e}^{2\pi\mathrm{i}\mathbf{z}^{\top}\boldsymbol{\xi}} \frac{g(\boldsymbol{\xi})}{p(\boldsymbol{\xi})}\right] \leq c^2 - f(\mathbf{z})^2.
    \end{equation}
\end{lemma}

\begin{proof}
    Recall that for a complex random variable $Z$, its variance is defined as
    \begin{equation}\label{eq:var-definition}
        \mathrm{Var}[Z] = \mathbb E \left[(Z-\mathbb E \left[Z\right])(Z-\mathbb E \left[Z\right])^*\right] = \mathbb E \left[ZZ^*\right] - \mathbb E \left[Z\right] \mathbb E \left[Z\right]^*,
    \end{equation}
    where $^*$ denotes the conjugate.

    Straightforward calculation gives
    \begin{align}
        \mathbb E \left[\left(\mathrm{e}^{2\pi\mathrm{i}\mathbf{z}^{\top}\boldsymbol{\xi}} \frac{g(\boldsymbol{\xi})}{p(\boldsymbol{\xi})}\right)\left(\mathrm{e}^{2\pi\mathrm{i}\mathbf{z}^{\top}\boldsymbol{\xi}} \frac{g(\boldsymbol{\xi})}{p(\boldsymbol{\xi})}\right)^*\right] =  & \mathbb E \left[\mathrm{e}^{2\pi\mathrm{i}\mathbf{z}^{\top}\boldsymbol{\xi}} \frac{g(\boldsymbol{\xi})}{p(\boldsymbol{\xi})} \cdot \mathrm{e}^{-2\pi\mathrm{i}\mathbf{z}^{\top}\boldsymbol{\xi}} \frac{g(\boldsymbol{\xi})^*}{p(\boldsymbol{\xi})}\right] \\
        = & \mathbb E \left[\frac{|g(\boldsymbol{\xi})|^2}{p(\boldsymbol{\xi})^2}\right]\leq c^2.
    \end{align}

    Besides, Eq. (\ref{eq:fourier}) implies that 
    \begin{equation}
        \mathbb E \left[\mathrm{e}^{2\pi\mathrm{i}\mathbf{z}^{\top}\boldsymbol{\xi}} \frac{g(\boldsymbol{\xi})}{p(\boldsymbol{\xi})}\right] = f(\mathbf{z}).
    \end{equation}

    Plugging the above two results into Eq. (\ref{eq:var-definition}) yields Eq. (\ref{eq:var-flt}) and hence concludes the proof.
\end{proof}

\begin{theorem}[Variance of RPE approximation]
    \label{appthm:var-flt}
    Under the assumption of Lemma \ref{applemma:var-flt}, for any $\mathbf{x}, \mathbf{y}\in\mathbb{R}^{\ell}$, the variance of the approximation given by $\varphi(\mathbf{x})^{\top}\psi(\mathbf{y})$ in Theorem \ref{appthm:rpe-decompose} satisfies
    \begin{equation}\label{eq:var-flt-corollary}
         \mathrm{Var}[\varphi(\mathbf{x})^{\top}\psi(\mathbf{y})]\leq \frac{c^2 - f(\mathbf{x}-\mathbf{y})^2}{r}.
    \end{equation}
\end{theorem}

\begin{proof}
    Note that
    \begin{equation}
        \varphi(\mathbf{x})^{\top}\psi(\mathbf{y}) = \frac{1}{r}\sum_{t=1}^r
        \mathrm{e}^{2\pi\mathrm{i}(\mathbf{x}-\mathbf{y})^{\top}\boldsymbol{\xi}_t} \frac{g(\boldsymbol{\xi}_t)}{p(\boldsymbol{\xi}_t)}, 
    \end{equation}
    where the random features $\boldsymbol{\xi}_1, \cdots, \boldsymbol{\xi}_r$ are $r$ i.i.d. samples from the distribution $p$.

    Setting $\mathbf{z}=\mathbf{x}-\mathbf{y}$ in Eq. (\ref{eq:var-flt}) and considering $r$ i.i.d. samples immediately yield Eq. (\ref{eq:var-flt-corollary}).
\end{proof}

\subsection{Uniform convergence and sample complexity of RPE approximation}

\begin{theorem}[Uniform convergence and sample complexity for approximation]\label{appthm:sample-complexity-flt}
    \footnote{This is Theorem {\color{blue} 4.2} in the main paper.}
    Given $L$ vectors $\mathbf{r}_{1},...,\mathbf{r}_{L} \in \mathbb{R}^{\ell}$, define the RPE attention mask $\mathbf{N} = [f(\mathbf{r}_{i}-\mathbf{r}_{j})]_{i,j\in [L]}$.   
    Assume that $c= \||g(\mathbf{x})|/p(\mathbf{x})\|_{\infty}$, where $g$ is the Fourier Transform of $f$ and $p$ is some probability density function over $\mathbb{R}^{\ell}$.
    
    For any $\varepsilon, \delta>0$, if the number of random features $r=\Theta \left(\frac{c^2}{\varepsilon^2}\log\frac{L}{\delta}\right)$, then \FLT 's RPE approximator $\widehat{\mathbf{N}}$ satisfies 
    $$\mathbb{P}\left(\|\mathbf{N}-\widehat{\mathbf{N}}\|_{\max}\leq \varepsilon \right)>1-\delta,$$
    where $\|\cdot\|_{\max}$ denotes the max norm of a matrix.
\end{theorem}

\begin{proof}
    For any $i, j\in\{1, \cdots, L\}$, the assumption $c= \||g(\mathbf{x})|/p(\mathbf{x})\|_{\infty}$ implies that almost surely
    \begin{equation}\label{eq:bounded-rv}
        \left|\mathrm{e}^{2\pi\mathrm{i}(\mathbf{r}_i-\mathbf{r}_j)^{\top}\boldsymbol{\xi}_t} \frac{g(\boldsymbol{\xi}_t)}{p(\boldsymbol{\xi}_t)}\right| \leq c \qquad (\forall~t\in\{1, \cdots, r\}),
    \end{equation}
    where $\boldsymbol{\xi}_1, \cdots, \boldsymbol{\xi}_r$ denote the $r$ random features from the distribution $p$.

    Note that Eq. (\ref{eq:bounded-rv}) implies that both the real part and imaginary part of the estimated RPE is bounded. Applying Hoeffding Inequality (to the real part and imaginary part) yields
    \begin{equation}
        \mathbb{P}\left(
        \left|\frac{1}{t}\sum_{t=1}^r
        \mathrm{e}^{2\pi\mathrm{i}(\mathbf{r}_i-\mathbf{r}_j)^{\top}\boldsymbol{\xi}_t} \frac{g(\boldsymbol{\xi}_t)}{p(\boldsymbol{\xi}_t)} - f(\mathbf{r}_i-\mathbf{r}_j)\right|>\varepsilon
        \right)<4\mathrm{e}^{-\frac{r\varepsilon^2}{4c^2}}.
    \end{equation}

    By union bound over $i, j\in\{1, \cdots, L\}$, we have
    \begin{equation}
        \mathbb{P}\left(\exists~i,j\in\{1, \cdots, L\}, \text{ s.t. }
        \left|\frac{1}{t}\sum_{t=1}^r
        \mathrm{e}^{2\pi\mathrm{i}(\mathbf{r}_i-\mathbf{r}_j)^{\top}\boldsymbol{\xi}_t} \frac{g(\boldsymbol{\xi}_t)}{p(\boldsymbol{\xi}_t)} - f(\mathbf{r}_i-\mathbf{r}_j)\right|>\varepsilon
        \right)<4L^2 \mathrm{e}^{-\frac{r\varepsilon^2}{4c^2}}.
    \end{equation}
    
    Equivalently, with probability at least $4L^2 \mathrm{e}^{-\frac{r\varepsilon^2}{4c^2}}$, we have
    \begin{align}
        & \left|\frac{1}{r}\sum_{t=1}^r
        \mathrm{e}^{2\pi\mathrm{i}(\mathbf{r}_i-\mathbf{r}_j)^{\top}\boldsymbol{\xi}_t} \frac{g(\boldsymbol{\xi}_t)}{p(\boldsymbol{\xi}_t)} - f(\mathbf{r}_i-\mathbf{r}_j)\right|>\varepsilon
        \qquad (\forall~i,j\in\{1, \cdots, L\})\\
        \Rightarrow\qquad  & \|\mathbf{N}-\widehat{\mathbf{N}}\|_{\max}\leq \varepsilon.
    \end{align}

    Set 
    \begin{equation}
        r=\frac{4c^2}{\varepsilon^2}\log\frac{4L^2}{\delta}=\Theta \left(\frac{c^2}{\varepsilon^2}\log\frac{L}{\delta}\right).
    \end{equation}

    Then we have
    \begin{equation}
        \mathbb{P}\left(
        \|\mathbf{N}-\widehat{\mathbf{N}}\|_{\max}\leq \varepsilon
        \right)>1-4L^2 \mathrm{e}^{-\frac{r\varepsilon^2}{4c^2}}=1-\delta,
    \end{equation}
    which concludes the proof.
\end{proof}

\subsection{Discussions on the theoretical results}

Theorem \ref{appthm:sample-complexity-flt} implies that \FLT 's estimated RPE mask $\widehat{\mathbf{N}}$ can approximate the true RPE mask $\mathbf{N}$ up to arbitrary precision with high probability. Besides, note that the constant $c$ can be viewed as fixed for a pretrained model which does not depend on $L$. Thus, in order to obtain an arbitrarily accurate RPE approximator, the required number of random features only scales \textit{logarithmically} with $L$. This property is particularly appealing because it indicates \FLT can remain accurate in the long sequence regimes while accelerating powerful RPE-enhanced attention.

Theorems \ref{appthm:var-flt} \& \ref{appthm:sample-complexity-flt} also provide insights on finding the optimal $p$. Note that the variance and the sample complexity both scale with $c= \||g(\mathbf{x})|/p(\mathbf{x})\|_{\infty}$, and lower $c$ can potentially leads to better approximation. Specifically, choosing $p(\mathbf{x}) \propto |g(\mathbf{x})|$ minimizes $c$ under the constraint of $p$ being a probability density function. The shift-invariant kernels RPE indeed satisfies this property, and is optimal in terms of the approximation sample complexity. 

Another simple yet effective approach is to parameterize $p$ as a Gaussian distribution with \textit{learnable} means and variances. The optimization procedure can search for the optimal $p$ in the class of Gaussian distributions to obtain good RPE approximation. This technique turns out to be helpful for the experiment on molecular property prediction (Sec. {\color{blue} 5.3} in the main paper).

Finally, we point out that the $\log L$ factor in the sample complexity bound in Theorems \ref{appthm:sample-complexity-flt} is introduced for technical reasons: the convergence analysis is conducted for the random features applying exponential mapping which is not bounded.\footnote{For random features applying trigonometric functions and thus could leverage $\varepsilon$-net trick combined with strong Lipschitz function argument. An analogous result can be found in \cite{choromanski}} That being said, this logarithmic factor is still negligible (as opposed to polynoimal dependency).

\section{DETAILED EXPERIMENT SETTINGS}
\label{app:exp}

All tested Transformer variants were trained and tested on a TPU pods containing $4$ \href{https://cloud.google.com/tpu/docs/system-architecture-tpu-vm#:~:text=TPU%20v3%20configurations%20provide%20significant,bound%20on%20TPU%20v3%20configurations.}{TPU v3 chips} with \href{https://jax.readthedocs.io/en/latest/notebooks/quickstart.html}{JAX} and on GPUs (V100).

\subsection{Language modeling} 
\label{app:lm}
In this experiment, we study \FLT~with two RPE variants, Gaussian mixture RPE and local RPE. The detailed descriptions of baselines have been provided in the main paper.

All the tested models are efficient Transformers based on kernelized low-rank attention, with 6 decoder layers. In each layer, there are 8 attention heads. The hidden dimension is set to 512. The dimension of the feed-forward sub-layer is set to 2048. The feature map dimension $m$ is set to 64 in the low-rank approximation of the attention matrix. For our \FLT~models, the number of random features for RPE $r$ is set to 32. We use the validation perplexity as the evaluation metric: lower perplexity indicates better performances.

Following existing works \cite{peng2021random, rpe-performers}, the sequence length is set to 512 during both training and evaluation. All  models are trained \textit{without} access to the context from previous mini-batches for a fair comparison. The dropout ratio and  weight decay are set to 0.1 and 0.01, respectively. The batch size is set to 64. We use $\mathrm{Adam}$ as the optimizer, and set its hyperparameter $\varepsilon$ to $1\mathrm{e}-6$ and $(\beta_1,\beta_2)$ to (0.9, 0.98). The model is trained for 150k steps with a 6k-step warm-up stage followed by an inverse square-root learning rate scheduler, with the peak learning rate set to $2\mathrm{e}-3$. 

For the \FLT~variant with Gaussian mixture RPE, the FT of the RPE function, i.e., the function $g$, is parameterized as Eq. (\ref{eq:gaussian-mixture-rpe}):
\begin{equation*}
g(\boldsymbol{\xi}) = \sum_{t=1}^{T} w_{t} \exp\left(-\frac{\|\boldsymbol{\xi}-\boldsymbol{\mu}_{t}\|^{2}}{2 \sigma_{t}^{2}}\right).
\end{equation*}

For the \FLT~variant with local RPE, the function $g$ is parameterized as 
\begin{equation}
g(\xi) = \sum_{t=1}^T w_t \cdot \frac{\sin(2\pi v_t \xi)}{\pi \xi},
\end{equation}
where $w_1, \cdots, w_T$ and $v_1, \cdots, v_T$ are learnable parameters and $T$ is a pre-defined hyper-parameter. In this case, the underlying implict RPE function $f$ is
\begin{equation}
f(\Delta r) = \sum_{t=1}^T w_t \cdot \mathbb{I}[|\Delta r| \leq v_t].
\end{equation}

For both \FLT~variants, the RPE masks are different in different attention heads, but are \textit{shared} across different layers. The random features $\xi_1, \cdots, \xi_r$ are sampled from the standard Gaussian distribution.

\subsection{Image classification} 
\label{app:image-exp}
Table \ref{tab:img-data-details} presents the basic statistics of the datasets used in the image classification experiements.

\begin{table}[!ht]
    \begin{center}
    \caption{Details of the datasets used in image classification tasks with the \textrm{FourierLearner-Transformer}.}\label{tab:img-data-details}
    \vspace{2mm}
    \begin{tabular}{@{}lccc@{}}
        \toprule
        \textbf{Dataset name} & \# of classes & Training set size & Test set size \\
        \midrule
        ImageNet2012 \cite{imagenet} & 1K & 1.2M & 100K  \\
        Places365 \cite{places} & 365 & 1.8M & 328K  \\
        Fashion-MNIST \cite{fashionmnist} & 10 & 60K & 10K  \\
        \bottomrule
    \end{tabular}
    \end{center}
\end{table}


All tested models consist of 12 layers with 12 attention heads in each layer. The dimension of the feed-forward sub-layer is set to 3072. 
In our \FLT, we use learnable $\mathrm{ReLU}$ as the feature map for kernelized linear attention. In particular, the feature map is $\phi: \mathbf{x}\mapsto \mathbf{Wx}$ where $\mathbf{W}$ is a learnable matrix. For all the models, we used a dropout rate of 0.1 and no attention dropout. We applied the $\mathrm{Adam}$ optimizer with weight decay equal to 0.05 and a standard batch size of 4096. All Transformers were trained on TPU architectures until convergence.


\begin{table}[ht]
\centering
\caption{Hyperparameters for Image Classification.}
\label{tab:vitpretrain}
\vspace{2mm}
\begin{tabular}{ccc}
    \toprule
    Parameter & & Value  \\ \midrule
    Batch size & & $4096$  \\
    Optimizer & & AdamW \\
    Base Learning rate & & $1.5\mathrm{e}-4$  \\
    Weight decay & & 0.05 \\
    Optimizer momentum & & $(\beta_1, \beta_2)=( 0.9, 0.95 )$  \\
    Learning rate schedule & & cosine decay \\
    Warm up epochs & & 40 \\
    Augmentation & & RandomResizedCrop \\
    Compute resources & & $8 \times 8$ TPUv3 \\ \bottomrule
\end{tabular}
\end{table}

\subsection{Molecular property prediction} 
\label{app:oc20}
We adopt most of the training strategies of 3D-Graphormer \cite{shi2022benchmarking}. 
Specifically, we trained a regular Performer with 12 layers and two \FLT~with 10 and 12 layers respectively. 
Following existing works \cite{jumper2021highly, shi2022benchmarking}, model outputs are repeatedly fed to the model for four times. In each layer, there are 48 attention heads. The hidden dimension is set to 768. The dimension of the feed-forward sub-layer is set to 2048. The feature map dimension $m$ is set to 64 in the low-rank approximation of the attention matrix. For our \FLT~models, the number of random features for RPE $r$ is set to 16 and the number of Gaussian basis functions in RPE $T$ is set to 32. The random feature $\xi_i$ are sampled from Gaussian distribution $\mathcal{N}(0,\sigma_i^2 \mathbf I)$, where $\sigma_i$ is learnable.

We evaluate the performance of the tested models on the in-domain validation set, where the validation samples come from the same distribution as the training distribution. We use Mean Absolute Error (MAE) of the energies and the percentage of Energies within a Threshold (EwT) of the ground truth energy to evaluate the accuracy of the predicted energies. 

For all the models, the attention dropout ratio and the weight decay are set to 0.1 and 0.001, respectively. The batch size is set to 64. We use $\mathrm{Adam}$ as the optimizer, and set its hyperparameter $\varepsilon$ to $1\mathrm{e}-6$ and $(\beta_1,\beta_2)$ to (0.9, 0.98). 
The peak learning rate is set to $3\mathrm{e}-4$ with a 10K-step warm-up stage. After the warm-up stage, the learning rate decays linearly to zero. All the models are trained for 500k steps in total. 

\subsection{Learnable optimizers} 

In all the experiments, meta-training pipelines and the training recipes from \cite{jain2023mnemosyne} are applied.
Following \cite{jain2023mnemosyne}, all learnable models are used as memory units in the corresponding learnable optimizers and meta-trained in the exact same way on a small set of unrelated optimization tasks. Furthermore, all attention-based memory mechanisms are derived from \cite{jain2023mnemosyne}.

\section{VISUALIZATIONS}

\subsection{Local RPEs}
\label{sec:visual-local}

\begin{figure*}[ht!] 
\centering
  \includegraphics[width=0.95\linewidth]{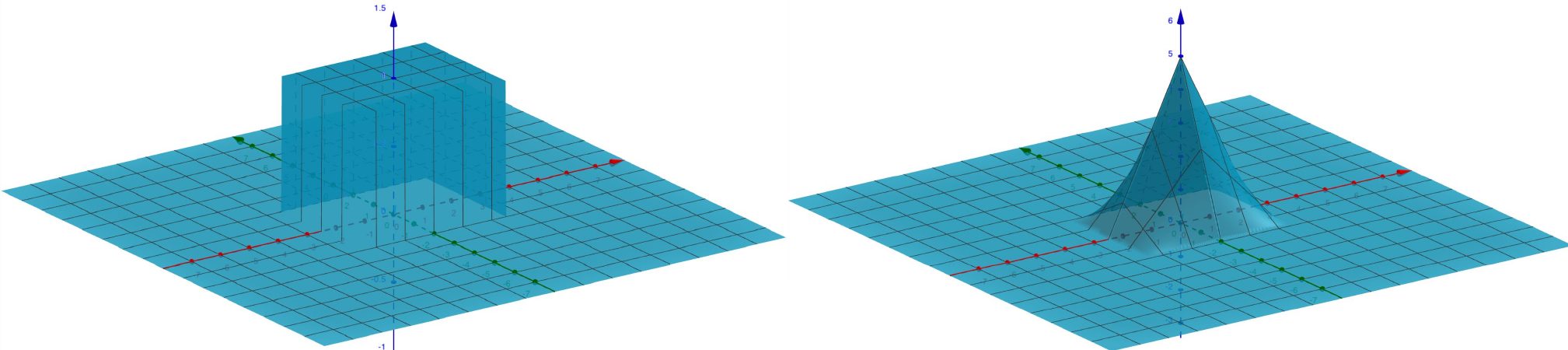}
\caption{\textbf{Examples of the local RPE mechanisms} discussed in Sec. \ref{sec:topology} and supported via \FLTs. Both examples are for tokens with positions described by two coordinates ($\ell=2$). The $x$ and $y$ coordinates encode the difference vector $\Delta \mathbf{r}=(\Delta r_1, \Delta r_2)^{\top}$. The $z$-coordinate provides the value of a function $f$. \textbf{Left:} (non-continuous) $f_{\mathbf{v},C}(\Delta\mathbf{r})=C \cdot \mathbb{I}_{\{|\Delta r_{1}| \leq v_{1}\}}\mathbb{I}_{\{|\Delta r_{2}| \leq v_{2}\}}$
for some $\mathbf{v}=(v_{1},v_{2})^{\top}$. \textbf{Right:} (continuous) $f_{\mathbf{v}}(\Delta \mathbf{r})=\mathbb{I}_{\{|\Delta r_{1}| \leq v_{1}\}}\mathbb{I}_{\{|\Delta r_{2}| \leq v_{2}\}}(-|\Delta r_{1}| + v_{1})(-|\Delta r_{2}| + v_{2})$. Both local RPE functions vanish outside the bounded region.
}
\label{fig:local}
\end{figure*}

In Fig. \ref{fig:local}, we visualize the shape of local RPEs that can be modeled with \FLTs~via Fourier Transform.

\subsection{Attention matrices}
\label{sec:attn-mat}
\begin{figure*}[ht!] 
    \centering
    \includegraphics[width=0.32\linewidth]{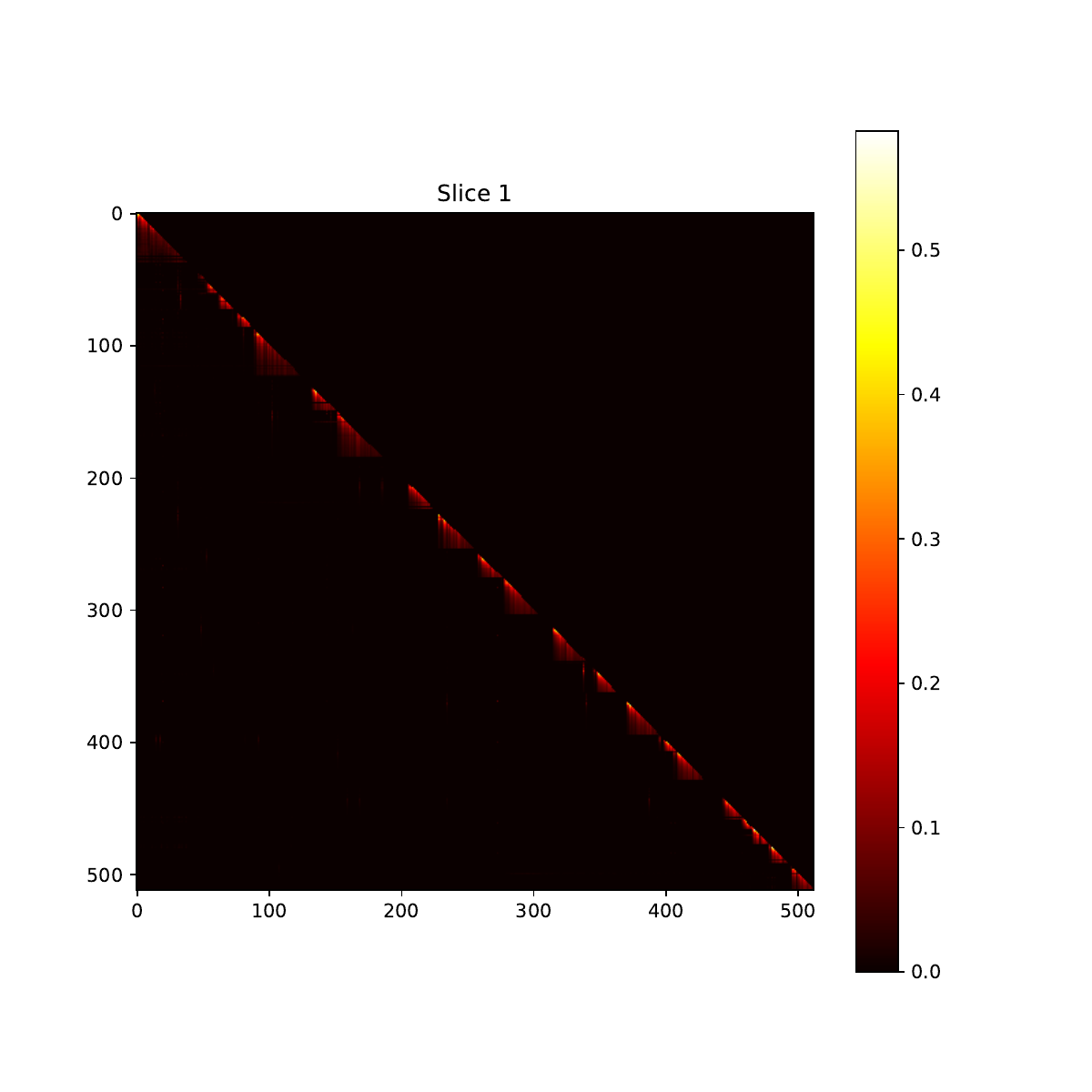}
    \includegraphics[width=0.32\linewidth]{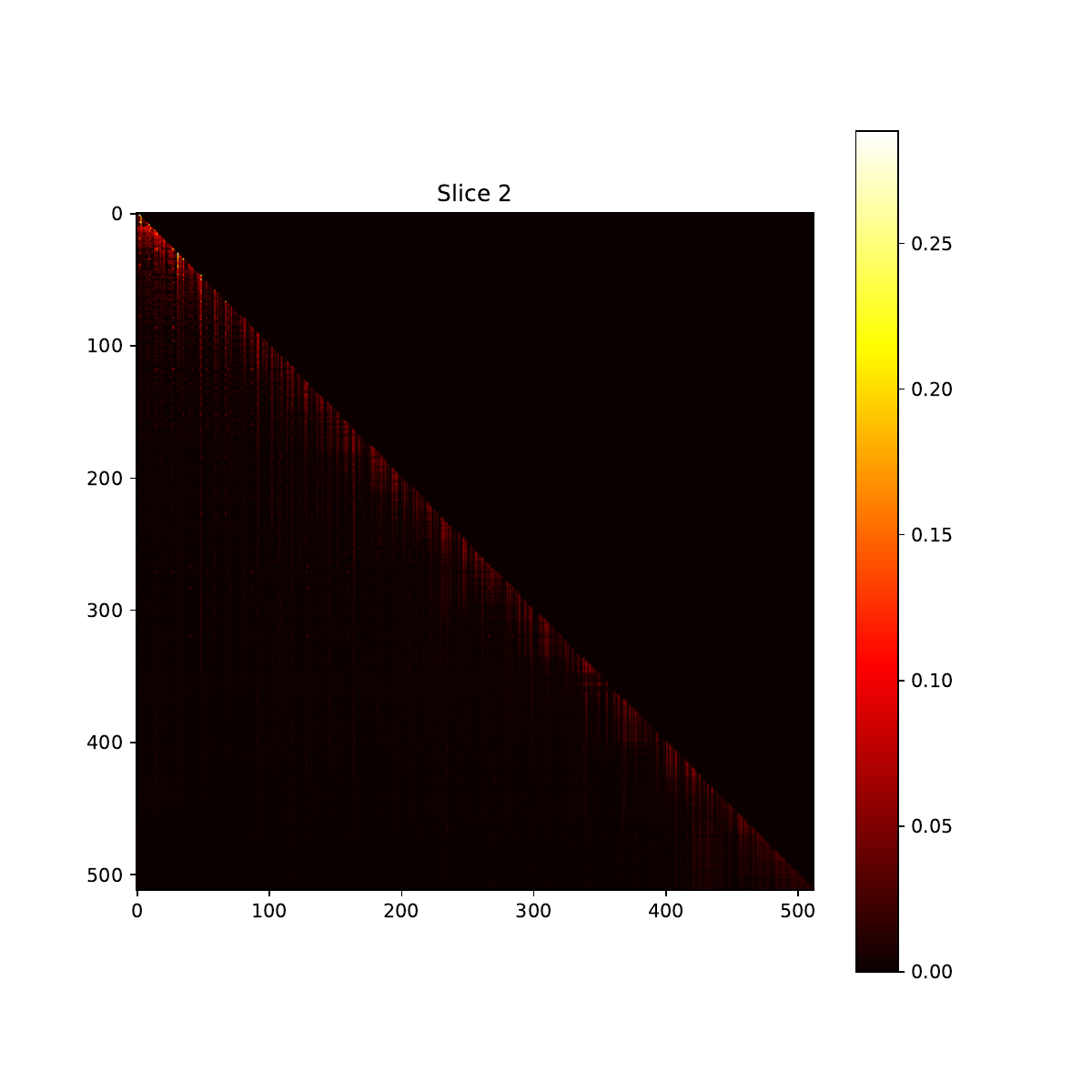}
    \includegraphics[width=0.32\linewidth]{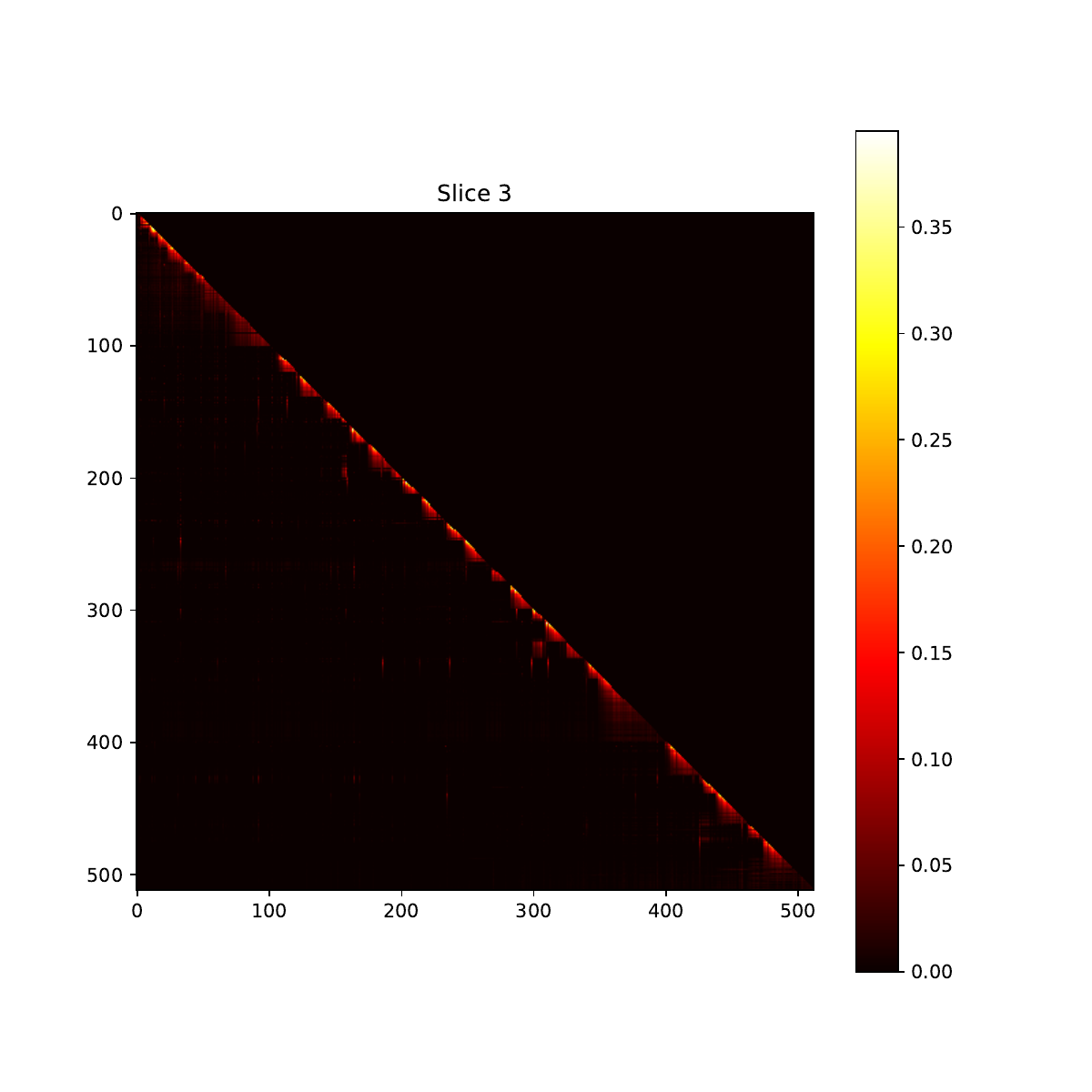}\\
    \vspace{-9mm}
    \includegraphics[width=0.32\linewidth]{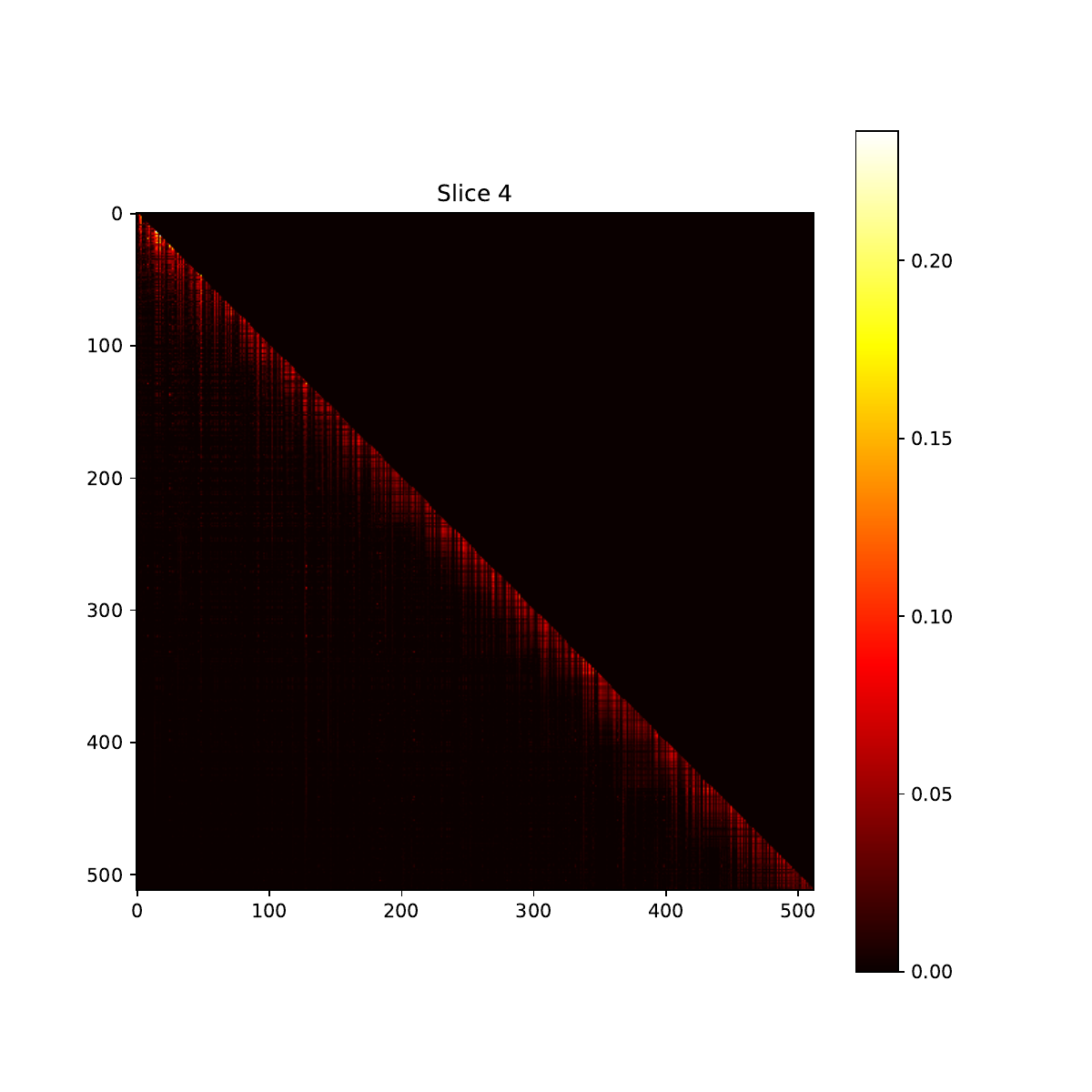}
    \includegraphics[width=0.32\linewidth]{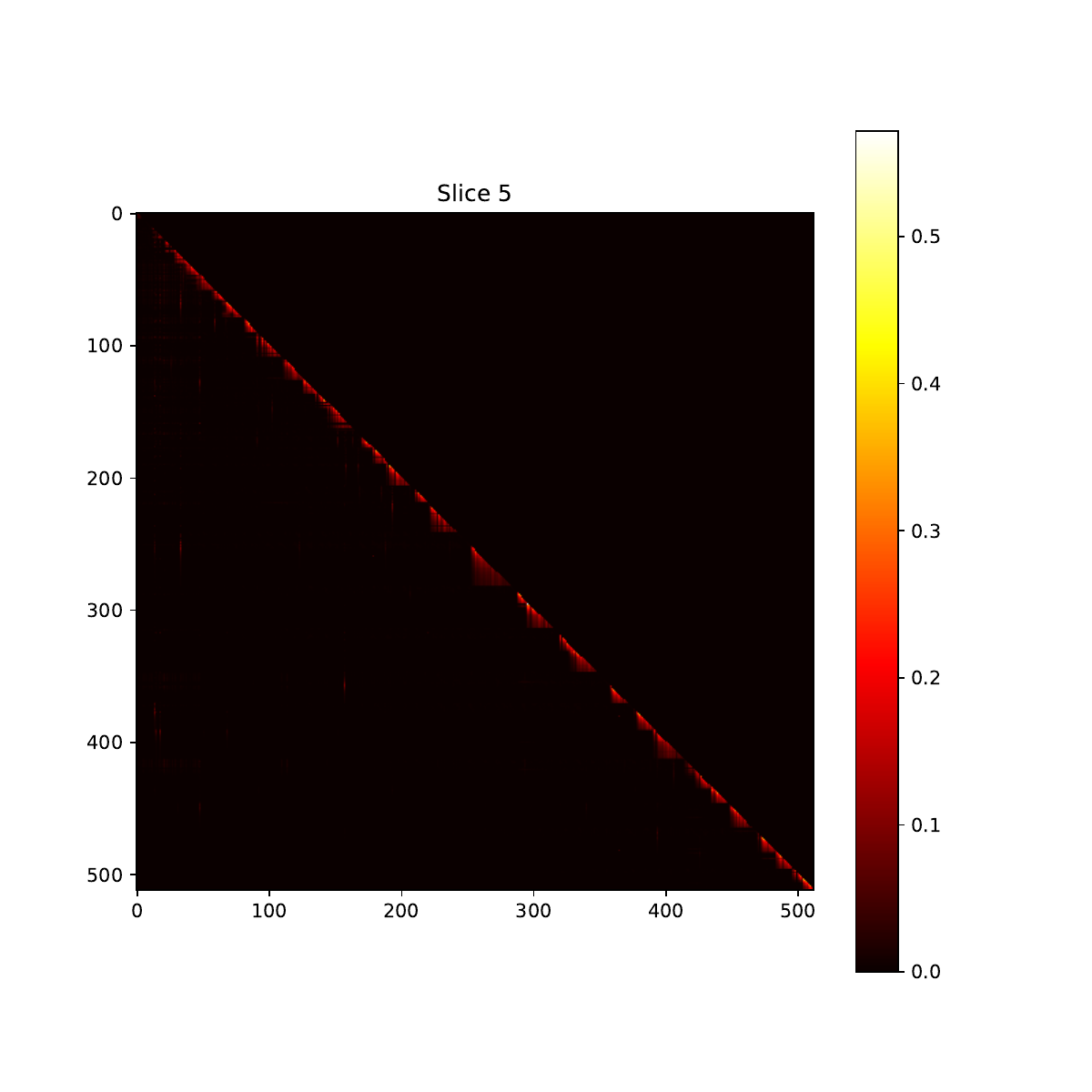}
    \includegraphics[width=0.32\linewidth]{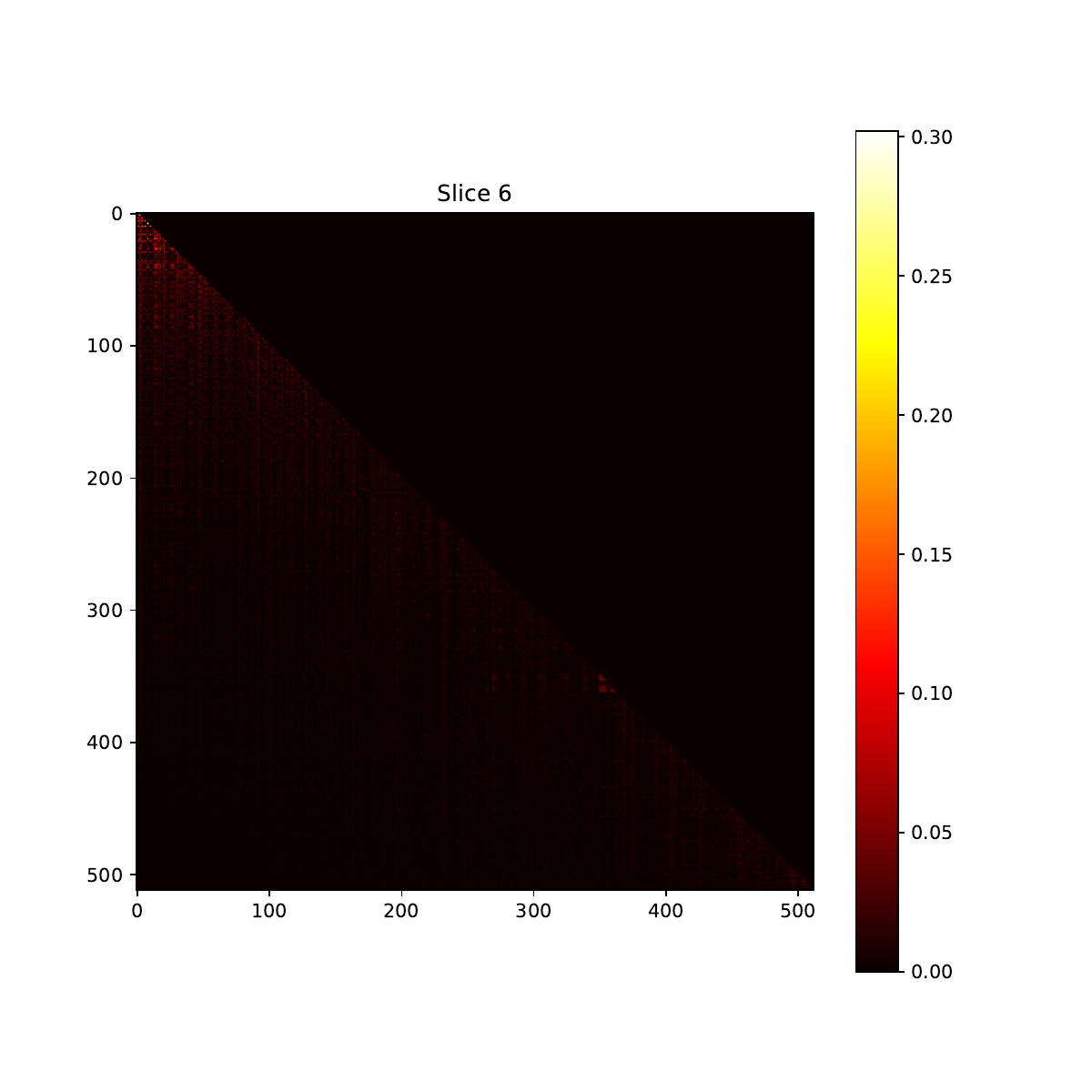}\\
    \vspace{-9mm}
    \includegraphics[width=0.32\linewidth]{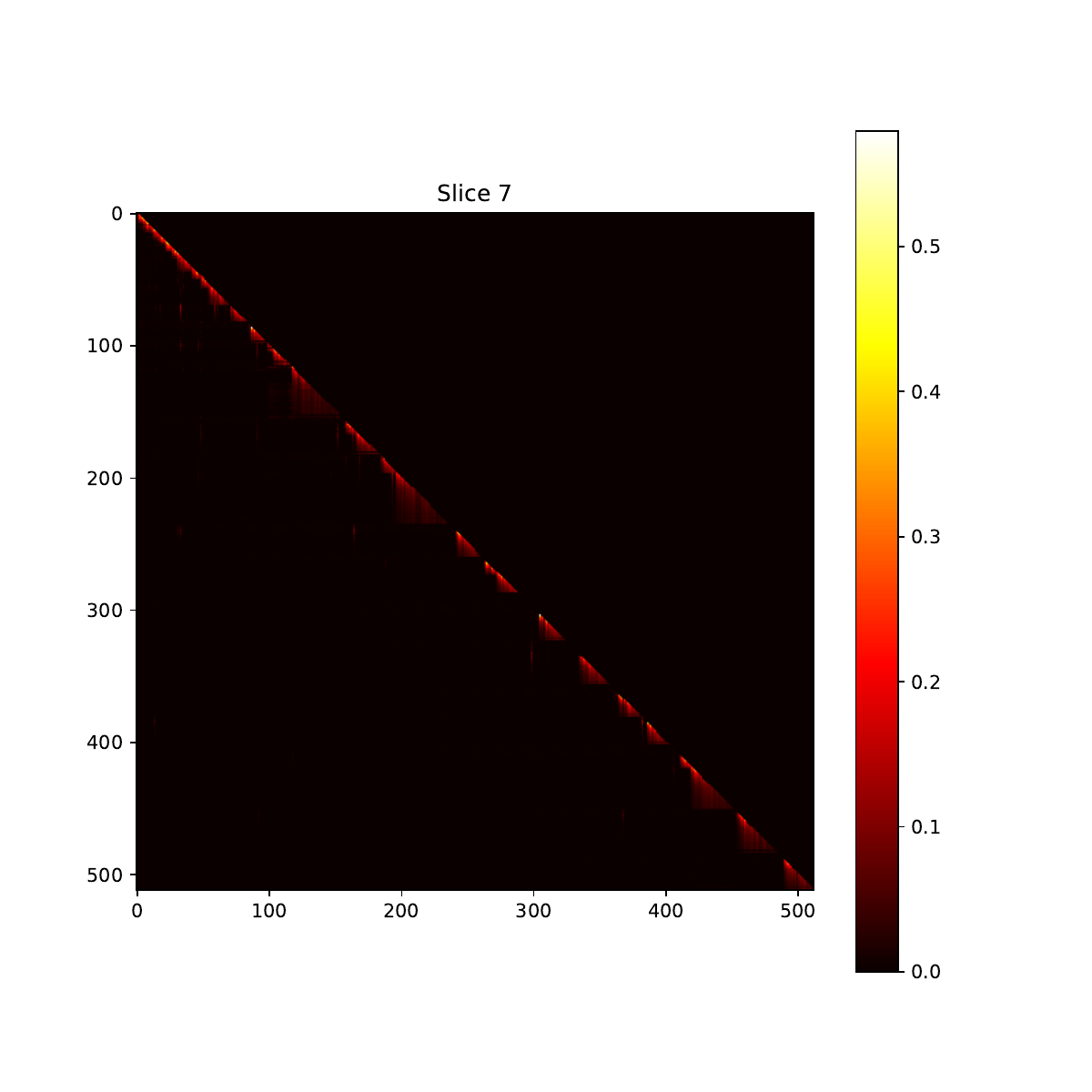}
    \includegraphics[width=0.32\linewidth]{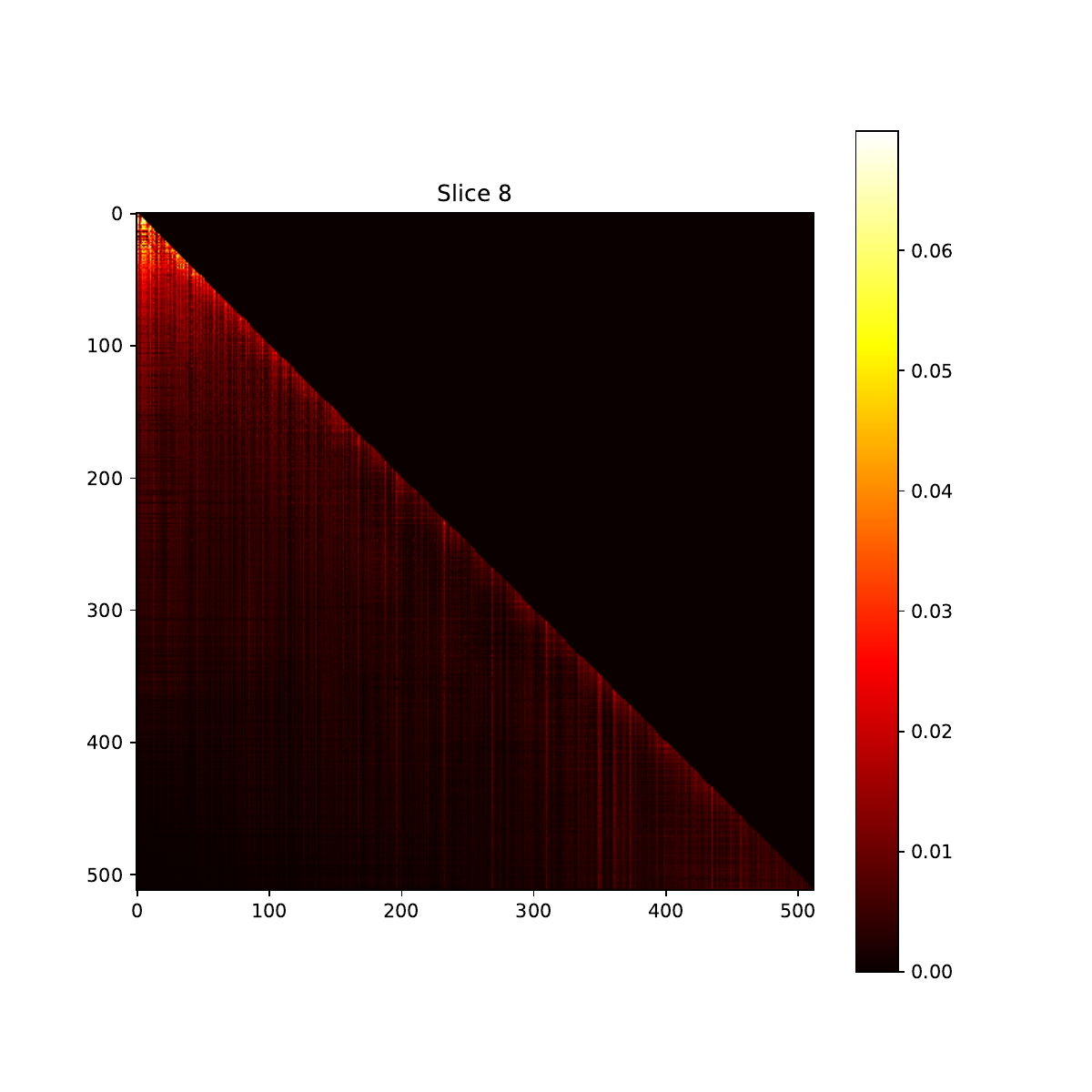}
    \vspace{-5mm}
\caption{\textbf{Attention matrix visualizations} of 8 attention heads in the first layer of \FLT~with local RPEs.}
\label{fig:attn-mat}
\end{figure*}

In Fig. \ref{fig:attn-mat}, we visualize the attention matrices of an \FLT~model trained on WikiText-103 language modeling. In particular, we feed one sequence in the training set as the input to the model and visualize the attention matrices of the 8 attention heads in the first layer. It can be seen that some attention heads pay more attention to nearby tokens, while others shows global attention patterns. The average attention probability over the most distant/nearby 10\% tokens is 0.068/0.279 respectively. This result shows that \FLT~ learns locality bias in language while maintaining the advantage to capture global contexts and leverage information in distant tokens.

\end{document}